\documentclass[twocolumn]{svjour3}
\smartqed 

\usepackage{cite}
\usepackage{amsmath,amssymb,amsfonts}
\usepackage{algorithmic}
\usepackage{adjustbox}
\usepackage{graphicx}
\usepackage{textcomp}
\usepackage{hyperref}
\usepackage{subcaption}
\usepackage{systeme}
\usepackage{tabularx}
\usepackage{placeins}
\usepackage[usestackEOL]{stackengine}
\usepackage[table,xcdraw]{xcolor}
\usepackage{color}
\usepackage{letltxmacro}
\usepackage[affil-it]{authblk}
\DeclareMathOperator*{\argmax}{arg\,max}
\DeclareMathOperator*{\argmin}{arg\,min}

\begin{document}


\title{Comparison of modern open-source Visual SLAM approaches}


\author{
    Dinar Sharafutdinov \and
    Mark Griguletskii \and
    Pavel Kopanev \and
    Mikhail Kurenkov \and
    Gonzalo Ferrer \and 
    Aleksey Burkov \and
    Aleksei Gonnochenko \and Dzmitry Tsetserukou 
}


\institute{
Skolkovo Institute of Science and Technology:\\
    Dinar Sharafutdinov \email{dinar.sharafutdinov@skoltech.ru},
    Mark Griguletskii \email{mark.griguletskii@skoltech.ru},
    Pavel Kopanev \email{pavel.kopanev@skoltech.ru} \and 
    Mikhail Kurenkov \email{mikhail.kurenkov@skoltech.ru},
    Gonzalo Ferrer \email{G.Ferrer@skoltech.ru},
    Dzmitry Tsetserukou \email{D.Tsetserukou@skoltech.ru}\\
    Sberbank Robotics Lab:\\
    Aleksey Burkov
    \email{burkov.a.m@sberbank.ru},
    Aleksei Gonnochenko
    \email{gonnochenko.a.s@sberbank.ru}
}

\date{Received: date / Accepted: date}

\maketitle



\begin{abstract}\hspace{0.5cm}
Simultaneous localization and mapping (SLAM) is one of the fundamental areas of research in robotics and environment reconstruction. State-of-the-art solutions have advanced significantly in terms of mapping quality, localization accuracy and robustness. It becomes possible due to modern stable solvers in the back-end, efficient outlier rejection techniques and diversified front-end: unique features, topologically segmented landmarks, and high-quality sensors. Among the variety of open-source solutions, several promising approaches provide results which are difficult to be reproduced on standard datasets, especially if there is no description for dataset adaptation. The goal of the article is to figure out, which techniques of robots' localization are the most promising for further use in related disciplines for engineers and researchers. The main contribution is a comparative analysis of state-of-the-art open-source Visual SLAM methods in terms of localization precision for versatile environments. The algorithms are assessed based on accuracy, computational performance, robustness and fault tolerance. Additionally, the survey and comparison of the datasets used for methods evaluation are provided as well as practical recommendations of usage scenarios for further research.
\end{abstract}




\section{Introduction}

Simultaneous localization and mapping (SLAM) is an important task in computer robotics, computer vision and environment reconstruction. Robots need to estimate their current position and surrounding map during operation. SLAM algorithms require measurements between sequential positions of the robot (aka odometry) and between the robot and landmarks. Thus, the goal of SLAM is to reconstruct a consistent map of the environment and localize a robot on it. The problem becomes much more difficult in an unknown environment with no prior information.

There are many open-source solutions that have been developed over the last 20 years. However, it is a complex problem of selecting an open-source SLAM system among the options available. The algorithms are implemented differently and tested on diverse datasets. At the same time, a good common evaluation of SLAM systems is essential for robotics engineers. They must know weaknesses and strengths to properly use SLAM solutions in robotic systems.

There are several approaches to solve the visual SLAM problem. Main methods are feature-based \cite{orb} and direct methods \cite{engel2014lsd}. Feature-based methods use key points. In these methods, key points are saved to local submaps or key-frames and then a graph optimization approach is applied. Whereas direct-SLAM methods save all pixels of images on local maps and use photometric losses for measurement error estimations. Modern SLAM approaches are very different and usually highly modular. They consist of front-end and back-end. Moreover, some state-of-the-art methods solve the SLAM problem with dynamic objects. 

\begin{figure}[h]
\centering
    \begin{subfigure}{.49\textwidth}
        \centering
        \includegraphics[width=0.98\textwidth]{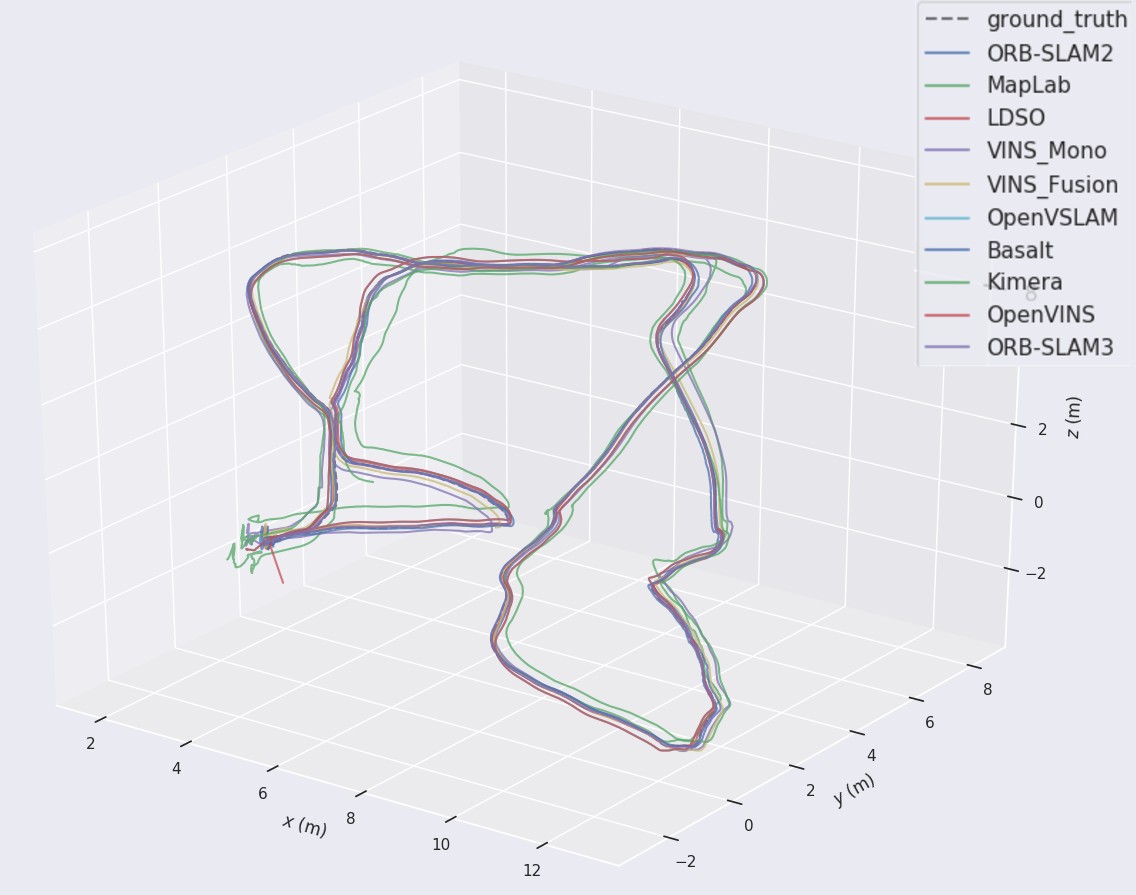}
        \caption{EuRoC MAV mh\_5}
    \end{subfigure}
    \begin{subfigure}{.49\textwidth}
        \centering
        \includegraphics[width=0.98\textwidth]{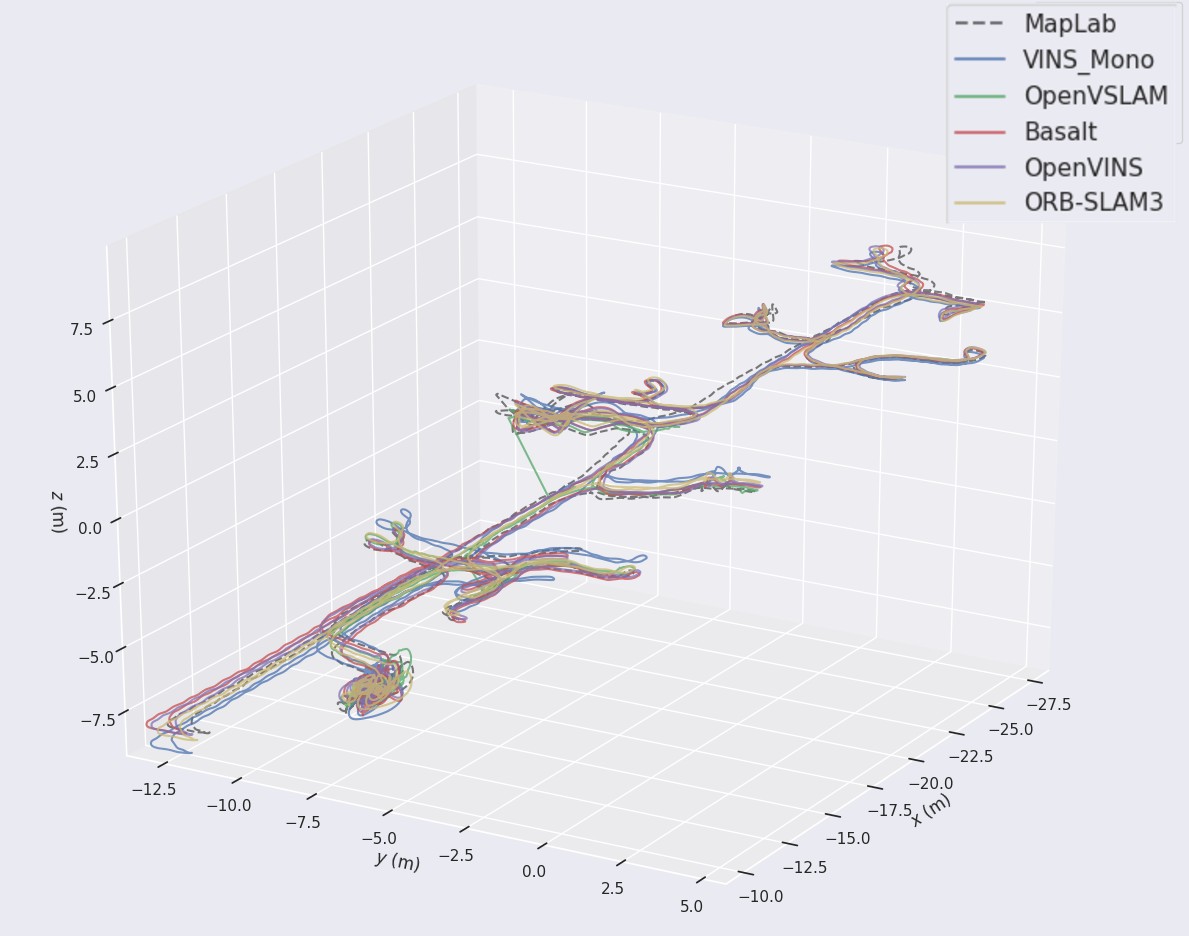}
        \caption{TUM VI corridor\_1}
    \end{subfigure}
    \caption{Examples of resulting trajectories.}
\label{fig_trajectories}
\end{figure}

During the last years, SLAM systems for different sensors apart from RGB cameras have been developed. For example, RGB-D (depth) and stereo cameras \cite{orb2} as well as IMU measurements\cite{vinsmono} or transformations between lidar point clouds\cite{NDT}. The combination of different information about the environment leads to a consistent solution to the SLAM problem. Thus, the approaches must be implemented and evaluated on the datasets with versatile sensors. This paper focuses on feature-based solutions: ORB-SLAM 2/3, MapLab, LDSO, VINS-Mono, VINS-Fusion, Open VSLAM, Basalt, Kimera, Open VINS, and DRE. Section 3.1 (Choice of algorithms) describes the choice of algorithms in detail. A full comparison of localization accuracy of different algorithms as well as memory and computational resources is represented in section 4 (Results). The practical aspects, description of experiments and datasets review could be found in section 3 (Experimental setup). The facilities of original algorithms such as availability of documentation, examples on popular datasets, the convenience of the interface, ability to change the parameters of algorithms and presence of Docker/ROS wrappers (table \ref{table:pract_compare} for more details) are qualitatively evaluated.

\begin{table}[h]
\centering
\caption{Practical facilities of algorithms.}
\label{table:pract_compare}
\resizebox{0.48\textwidth}{!}{
\begin{tabular}{|c|c|c|c|c|c|c|} 

\hline
 \bfseries{Framework}  & \bfseries{Documentation} & \bfseries{Support}  & \textbf \bfseries \Centerstack{\bfseries{Usage examples} \\ \bfseries{on popular datasets}}  & \bfseries{Docker} & \bfseries{ROS} \\ 
\hline

ORB-SLAM2 & github & - & + & - & + \\ 
\hline
MapLab & github & - & + & - & +\\
\hline
LDSO & github & + & + & - & -\\
\hline
VINS-Mono & github & - & + & + & +\\
\hline
VINS-Fusion & github & - & + & + & +\\ 
\hline
OpenVSLAM & web-page, github & + & + & + & +\\ 
\hline
Basalt & github & + & + & + & -\\
\hline
Kimera & github & + & EuRoC only & + & +\\
\hline
OpenVINS & web-page, github & + & + & - & +\\
\hline
ORB-SLAM3 & github & + & + & - & + \\ 
\hline
DRE & github & - & - & - & +\\
\hline

\end{tabular}}
\end{table}

Modern SLAM technologies achieve significant results \cite{ScaramuzzaFlyingVioComparison} but several challenges exist \cite{Cadena2016}. The main problems are long-term stability, scalability, sensor biases and miscalibrations, computation and memory requirements, dynamic objects, ambiguous scenes, and accuracy requirements. Since one of the most popular areas of SLAM usage is mobile robotics in the outdoor environment, critical issues are dynamic objects and map size. Thus, SLAM solutions should take these challenges into account and modern datasets and benchmarks need to be focused on these problems.

Nowadays, technical research and papers suffer a common issue with the reproducibility of their methods and results. This is called the “reproducibility crisis”. Frequently, it is very complicated to repeat the same results as described in paper or documentation. But this might be critical for researchers and engineers. Fast reproducibility allows testing and evaluating modern SLAM methods in their environments and choosing the appropriate setup for a particular task.

There is a number of datasets and benchmarks for SLAM evaluation. Some of them are well described in a relevant work \cite{Liu2021}. A dataset should reflect SLAM challenges, for example, dynamic objects, ambiguous scenes, and huge spaces.

The purpose of the current paper is to give a general introduction to algorithmic state-of-the-art solutions for visual SLAM methods and visual/visual-inertial odometry, to evaluate these algorithms on the same platform, compare the approaches on open datasets and give practical advice for SLAM researchers. Figure \ref{fig_trajectories} illustrates the trajectories for visualization purposes and alignment consistency.

The contribution of the article is a comparative analysis of modern Visual SLAM solutions which consists of results reproduction and multilateral comparison on various open datasets by evaluating localization accuracy, reproducibility, usability, needed computational resources and robustness to various scenarios. These aspects allow us to analyze the advantages and disadvantages of methods in different conditions and understand which of them are suitable for various situations. Additionally, the article provides a practical comparison and description of datasets used for methods evaluation as well as hints for researchers for choosing Visual SLAM methods supported by published GitHub repository\footnote{SLAM-Dockers \url{https://github.com/KopanevPavel/SLAM-Dockers}} with Dockers of the algorithms.
\section{Related work}
For localization and mapping, many sensors are used as input data. For example, 3D LIDARs, 2D LIDARs, depth cameras, stereo cameras, global and rolling shutters cameras. LIDARs are quite precise but very expensive. Moreover, it needs to emit light signals to estimate distance. Therefore, LIDARs are active sensors. One of the most beneficial sensors for robots is the rolling shutter camera. It is cheap, dense and passive. They do not need to emit light and only receive surrounding information. Global shutter cameras might be more convenient, but the price is higher. Stereo and depth cameras such as D435 are also appropriate and very popular. A comprehensive comparison of modern sensors for this problem was made by R. Singh and K. S. Nagla in \cite{SensorsComparison}. All of them can be used in SLAM pipelines, but they have different features which are important while solving the SLAM problem \ref{slam_equation}. The last is formulated as a maximization of conditional probability ($P$) of the robot`s states ($X$) and landmarks ($M$) given sensor measurements ($Z$) and information matrices ($\Omega$) which is equivalent to minimization of minus logarithm of $P$ in case of Gaussian structure of sensors` noises. It is equivalent to solving a non-linear least-squares problem or kernel-based ($\rho$) minimization for robustness in case of outliers in measurements.

\begin{equation} \label{slam_equation}
\begin{split}
        \hat{X},\hat{M}=\underset{X,M}\argmax\textbf{ }P\left(X,M|Z\right) \\
       \simeq{\underset{X,M}\argmin\textbf{ }\left\{-log\left(P\left(X,M|Z\right)\right)\right\}}\\
       \equiv{\underset{X,M}\argmin\textbf{ }\sum_{ij}{\rho(X_{ij},M_{ij},\Omega_{ij})}}
\end{split}
\end{equation}
Worth mentioning, that the main difference between SLAM and Visual/Visual-inertial odometry is that the last only updates the robot`s states ($X$) whereas pure SLAM additionally updates the map itself ($M$). This paper evaluates accuracy in terms of robot localization but not map quality.
Apart from visual sensors, inertial measurements are also being used as input: wheel odometry and IMU measurements are actively utilized in robotics.
SLAM has been extensively studied during the last decades. For example, Cadena et al. \cite{ Cadena2016} presented a good overview of modern SLAM technologies and the challenges which SLAM methods. According to another review \cite{Chen2020}, today SLAM is going into the spatial artificial intelligence age. This means that more deep learning techniques are used as a solution. 

\subsection{Sparse visual SLAM}
The history of feature-based SLAM or sparse visual SLAM began with the history of visual SLAM. One of the first successful approaches which solved the problem of simultaneous localization and mapping is a method presented by Davison et al. \cite{Davison2007}. In this work, the MonoSLAM algorithm had been introduced. It was able to find the camera position from the input images. The authors managed to speed up the solution of the SfM (Structure from Motion) problem by using methods based on probabilistic filters. The algorithm can obtain the position of point landmarks and the trajectory of the camera in real time.

There are several disadvantages of the MonoSLAM method. The first problem is the filtering approach which badly scales. The other disadvantage of the MonoSLAM algorithm was the necessity to update the trajectory and the feature positions sequentially in the same thread. 
To solve this problem, Klein et al. \cite{Klein2007} proposed the PTAM (Parallel Tracking and Mapping) algorithm
which achieves real-time performance on an algorithm that executes full batch updates.
Additionally, authors of the work \cite{Strasdat2010} showed that the solutions based on bundle adjustment \cite {Klein2007} are superior to filter-based ones.

\begin{figure*}[h]
    \centering
    \includegraphics[width=0.7\textwidth]{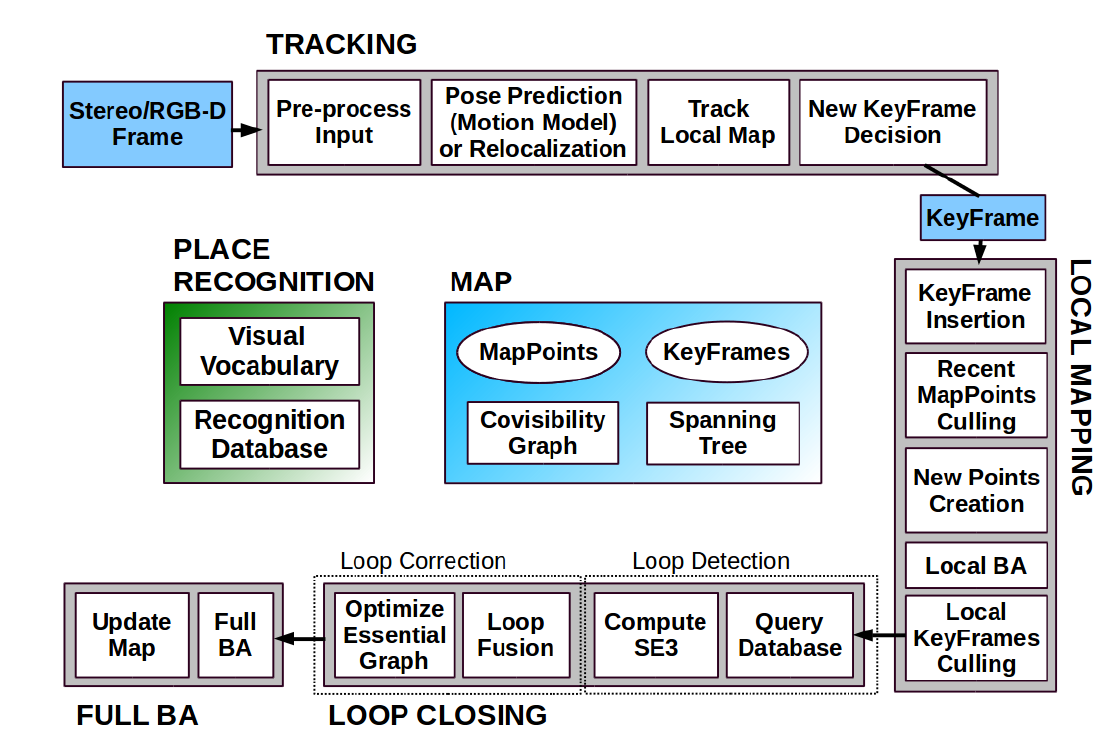}
    \caption{The scheme of the ORB-SLAM algorithm \cite{orb}.}
    \label{fig:orb_slam}
\end{figure*}

The key contribution in the field of solving the problem of visual SLAM was made by the ORB-SLAM algorithm which was described in this article \cite{orb}. The authors of the paper integrated all previous methods together in the same frame and combined the key elements of visual SLAM methods: Bags of Binary Words \cite{GalvezLopez2012} visual terrain recognition algorithm, ORB \cite{Rublee2011} algorithm for quick detection and description of key points and g2o \cite{Kummerle2011} optimizer. Figure \ref{fig:orb_slam} shows general scheme of the method. There are also other backend optimizers that researchers highly used. For example, GTSAM \cite{GTSAM} and Ceres \cite{ceressolver}. However, this article focuses more on different frontend parts of SLAM systems.

ORB-SLAM has three parallel processes. The first process is the construction of the local camera trajectory by matching the observed key points to the local map. The second process builds a local map and solves the local bundle adjustment problem. And the last process finds loop closures. Moreover, it can trigger a full optimization of the entire camera trajectory. The authors of ORB-SLAM also proposed a second version of their algorithm: ORB-SLAM2 \cite{orb2}. This algorithm allows using stereo and depth cameras.

ORB-SLAM method suffers from drifts of a scale. In order to reveal the geometric scale another source of information is often used, for example, an IMU sensor. The article \cite{Qin2017} introduces the VINS algorithm which uses the Kalman filter to merge sequential images and IMU data to calculate odometry. As a logical continuation of \cite{vinsfusion}, the authors started to use global optimization to relocate the camera.

The development of ORB-SLAM2 can be considered as SOFT-SLAM \cite{Cvisic2018}, the authors of this algorithm made the following changes:
\begin{itemize}
\item Instead of the more computationally complex bundle adjustment for localization, SOFT visual odometry is used, which achieves an error of about 0.8\% relative to the distance travelled.
\item The streams responsible for mapping and odometry are separated from each other, thus the visual odometry stream is not blocked by the thread responsible for mapping, which leads to a more stable processing time for incoming frames. In addition, unlike ORB-SLAM2, the algorithm is completely deterministic, i.e. always returns the same result for the same input.
\item SOFT key points are used for loop closure (similar to ORB-SLAM2), which makes the system highly efficient, simple and reliable while achieving sub-pixel precision. Despite the fact that SOFT key points are not invariant with respect to rotation, the authors show on public datasets that in practice, loop closure occurs often enough that this drawback has little effect on the final result.
\end{itemize}

It should also be noted that there is ORB-SLAM2-CNN \cite{Tateno2017}, which adds semantic information received from the neural network to the ORB-SLAM2 framework.

Another key point method worth mentioning is VINS-Mono \cite{Qin2018}. This method is focused on processing monocular data augmented with data from IMU.

The article \cite{Bustos2019} questions the visual SLAM approach based on global optimization of the entire trajectory and key point positions, that is, based on bundle adjustment. As an alternative, the authors of the article propose the L-infinity SLAM method. The essence of this method is that the robot first separately finds the camera rotations by averaging the relative rotations, after which it solves the optimization problem to find the three-dimensional position of the cameras and key points with known camera orientations. One can also use the rotation averaging described in the publication \cite {Bustos2019} for direct methods, and as additional restrictions for other ways to solve the visual SLAM problem.

\subsection{Dense visual SLAM}

Unlike methods for solving the problem of visual SLAM, which are based on key points, dense or direct methods do not use algorithmic features in their work. They use photometric error minimization for each pair of images to find the orientation and position of the camera in 3D space.

The first work that suggested using direct methods was DTAM \cite{Newcombe2011}. For each image, the DTAM algorithm generated a dense 3D map for each pixel. The next algorithm was KineticFusion \cite{Newcombe2011KineticFusion}, which uses a depth camera to build a dense 3D map. The authors use TSDF (truncated signed distance function) to describe each pixel, and ICP (iterative closest point) algorithm to map each depth image to a map.

The next milestone in the development of direct methods was the ElasticFusion \cite{Whelan2016} algorithm. This algorithm uses a direct representation of the surfaces of RGB-D cameras. For data fusion, this algorithm uses a non-rigid deformation model. Also in this algorithm, there is no global optimization of the graph of camera positions, and most of the steps take place on a graphics accelerator (GPU).

In parallel with the development of KineticFusion and ElasticFusion, two other teams were developing semi-dense visual odometry (SVO) \cite{engel2013semi, forster2014svo}. This algorithm can run in real-time on a robot processor (CPU). The SVO algorithm does not use all the pixels in the image, but only those that have a negligible gradient. The development of ideas from the publication \cite{engel2013semi} leads to the creation of a key work in this area - LSD-SLAM \cite{engel2014lsd}. LSD-SLAM uses $sim(3)$ - a metric to solve a problem with an uncertain scale, and also uses probabilistic inference to determine the error in constructing three-dimensional maps. The LSD-SLAM algorithm consists of three steps:
\begin{enumerate}
\item Get the image and determine the local offset relative to the current keyframe.
\item Update or create a new keyframe. If the current keyframe is updated, probabilistically merge the stereo depth data into the current keyframe. In case of creating a new keyframe, propagate the previous depth map to the new keyframe.
\item Updating the global map using position graph optimization. The edges in the graph are searched using the sim (3) metric, and the optimization is performed using the g2o library.
\end{enumerate} 

The LSD-SLAM development team continued the development of direct methods and proposed the DSO (direct sparse odometry) \cite{engel2017direct} method. Their method optimizes not only the photometric error but also the geometric error simultaneously. Also, instead of assuming smoothness, the authors use probabilistic pixel sampling. In LDSO \cite{Gao2018}, they added graph optimization to get a complete solution to the visual SLAM problem. In fact, LDSO is a combination of the ORB-SLAM and LSD-SLAM approaches.

Separately, it should be noted that the authors of DSO in 2019 published a modification of DSO \cite{schubert2019rolling}, adapted for processing frames obtained using a camera with a floating shutter. In addition to the photometric bundle adjustment, IMU data are also used as additional constraints for the optimization problem. Due to the fact that there is no selection of key points, this method (like DSO) can work not only with pixels representing edges and corners but with any pixels that have a sufficient gradient.

Many improvements to direct methods have also been proposed recently. For example, CodeSLAM \cite{Bloesch2018} suggests using autoencoders to train a more compact map representation for direct methods. Also, in the work KO-fusion \cite{houseago2019ko} it was proposed to use odometry and kinematics data obtained directly from the robot arm.


\subsection{Dynamic visual SLAM}
Usually, SLAM solutions suppose that scene is almost static or with a low level of dynamics. The simplest way of discarding outliers is to use RANSAC \cite{Fischler1987}. In reality, almost in every scenario, there will be lots of dynamic objects. Both indoor and outdoor scenes are full of people, animals, cars, pedestrians, bicycles and other dynamic objects. Therefore, it leads to changes in feature maps. This is especially important for loop closure detection. Additionally, in the case of a high level of dynamics, most slam solutions may show inconsistent results because of a lack of reliable features.    
There are several problems with dynamic objects. Firstly, the algorithms should detect the dynamic object and not use it for the trajectory estimation as well as for the mapping process. A possible solution here is to delete these objects from the image or replace them with a background. 
Popular tools for that are multi-view geometry\cite{dsslam}, semantic segmentation or detection neural networks\cite{dyna}, scene\cite{dyn} or optical flow and detection of background or foreground \cite{dre}. Secondly, there is a need of building a consistent map with or without dynamic objects. Solving the problem of lack of data due to the deletion of dynamic objects is an important aspect. Additional sensors, objects completion and reconstruction in combination with different map construction algorithms might help \cite{fusion4d}\cite{cofusion}. Dynamic SLAM algorithms often represent complex systems consisting of many parts and working not in real-time.

\section{Experimental setup}

There were several main goals in the experimental stage. First of all, the goal was to test the algorithms which are available to everyone. Only those SLAM solutions that have a link to the page with code and instructions were used. Secondly, it was decided to take one or few algorithms from each group of SLAM methods, instead of doing experiments with all the possible solutions.
It is crucial to see the results of algorithms that work with monocular, stereo and RGB-D cameras. Another essential part is understanding how additional data such as IMU or wheel encoder improves the results. The last point here is to try the solutions based on deep learning methods and with various front-end parts. Thirdly, there is the main use case that suits broad possible robotics tasks. Wheel robot with cameras and other sensors moves in an inside/outside environment. It moves among dynamic objects and features-less areas. The robot also visits the same places several times (loop closures), and the traversal could be relatively long (30 minutes). These factors affect the datasets that were used for tests. 

\subsection{Choice of algorithms}

\subsubsection{Sparse and dense SLAM} \hfill

ORB-SLAM2 \cite{orb2} was used as a baseline sparse SLAM solution. It is quite popular among researchers, and it proved efficiency in many scenarios. The second version is an expansion of the first one, hence ORB-SLAM1 \cite{orb} has not been used. Peculiarities of the algorithm were covered in the literature review section. 

Another algorithm here is an OpenVSLAM \cite{openvslam}. Generally, it is still ORB-SLAM but with additional features from ProSLAM and UcoSLAM. The main value of this solution is practical extensions such as support of the various type of camera models (perspective, fisheye, equirectangular), map saving and creation features, and finally convenient documentation and UI. 

The most modern approach which has been considered is the recently published ORB-SLAM3 \cite{orb3}. This sparse SLAM system is a natural development of the previous version of ORB-SLAM. Now it is a visual-inertial approach with an updated place recognition module (high-recall place recognition with geometrical and local consistency check) as well as with the support of pinhole and fish-eye camera models. Moreover, it allows localization in a multi-map setup.

Also, one of the dense SLAM methods have been chosen for comparison. LDSO \cite{Gao2018} is a graph-based dense SLAM system with loop closure detection. It is based on Direct Sparse Odometry (DSO) \cite{Engel-et-al-pami2018}. LDSO is robust and suitable for feature-poor environments utilizing any image pixel with sufficient intensity gradient. It uses a feature-based bag-of-words for loop closure detection and was validated on several open-source datasets.

\subsubsection{Visual-Inertial Odometry and SLAM}

A typical Visual-Inertial Odometry system implies the usage of a camera and IMU sensors. OpenVINS \cite{openvins} is Multi-State Constraint Kalman Filter (MSCKF) based VIO estimator. It uses IMU in the propagation step of the filter and camera data in the update step. In the current implementation OpenVINS also has a secondary loosely coupled loop closure thread based on VINS-Fusion \cite{vinsfusion}. Additionally, OpenVINS has an interface wrapper for exporting visual-inertial runs into the ViMap structure taken by maplab \cite{maplab}.  

VINS-Mono \cite{vinsmono} and VINS-Fusion is graph-based VIO approaches. VINS-Fusion is an extension of VINS-Mono and supports multiple visual-inertial sensor types (mono camera + IMU, stereo cameras + IMU, even stereo cameras only). Also, it supports global sensors like GPS and Barometer and has a global graph optimization module. Additionally, VINS-Fusion supports loop closure.

Basalt \cite{basalt} is other graph-based VIO approach. It has a lot of similarities with previously described VINS-Fusion (KLT feature tracking and Gauss-Newton non-linear optimization). But regarding map optimization, this approach utilizes ORB features.

Kimera \cite{Rosinol20icra-Kimera} is an open-source C ++ library that implements one way of solving SLAM problem in real-time. The modular structure includes: a VIO module with a GTSAM-based \cite{GTSAM} VIO approach, using  IMU-preintegration \cite{IMU_Preintegration}  and structureless vision factors \cite{Smart_Factors} for fast and accurate state assessment, a robust position graph optimizer (RPGO) for global trajectory estimation which adds a robustness layer that avoids  SLAM  failures due to perceptual aliasing,  and relieves the user from time-consuming parameter tuning, a lightweight 3D mesh module (Kimera-Mesher) for fast 3D mesh reconstruction \cite{3D_Mesh_Generation} and obstacle avoidance and a dense 3D semantic reconstruction module (Kimera-Semantics) that builds a  more-accurate global  3D  mesh using a  volumetric approach \cite{Voxblox}, and semantically annotates the 3D mesh using2D pixel-wise semantic segmentation based on deep learning methods.

\subsubsection{Dynamic SLAM} \hfill

In this section three algorithms have been considered: DynaSLAM \cite{dyna}, DynSLAM \cite{dyn} and DRE-SLAM \cite{dre}. All of them have theoretical potential and relatively popular implementations. DynaSLAM is a successor of ORB-SLAM2 with added front-end part that allows working in a dynamic environment. The solution consists of the segmentation of dynamic objects using CNN in monocular and stereo cases, and a combination of deep neural methods and multi-view geometry in the RGBD case. The important feature is that with DynaSLAM, it is possible to detect a priory dynamic objects, objects that may be static at the moment but the dynamic in essence. Pixel-wise semantic segmentation of potentially movable objects is possible with the help of Mask R-CNN. Then combining segmentation with multi-view geometry allows finding even static objects moved by dynamic ones (e.g. book in the hands of a person). After the deletion of dynamic objects, there are empty regions on images. To solve that, the authors use background inpainting by using information from previous views.

DynSLAM is a more complex solution that solves even more tasks for large-scale dynamic environments. Taking stereo images as the input, they compute depth maps (using ELAS or DispNet) and sparse scene flow (libviso2). Multi-task Network Cascades allow finding dynamic objects. Using obtained information, they do 3D object tracking and reconstruction. The final result is a static map without dynamic objects with the help of InfiniTAM. 
The last one, DRE-SLAM, is a bit different from the previous ones. Apart from image data, it also uses two-wheel encoders to handle the lack of features in dynamic scenes. It extracts ORB features from RGB-D images, uses YOLOv3 for dynamic object detection and then applies multiview constraint-based pixel culling. Loop closure detection is implemented through a bag-of-words approach. The final map is OctoMap which is constructed by fusing sub-OctoMaps. 

Although DynaSLAM and DynSLAM have developed and published solutions, the launch of the algorithms has not been successful. GitHub pages full of issues and implementations are not supported by authors, which makes solving varying problems very difficult. On the other hand, DRE-SLAM ran smoothly on ROS but it has examples only for data collected by authors. Comparison with other algorithms on popular datasets was difficult because of the lack of wheel encoder data in many of them. Nevertheless, the comparison has been performed with several other algorithms on the DRE dataset.

\begin{figure}[h]
\centering
    \begin{subfigure}{.42\textwidth}
        \centering
        \includegraphics[width=0.98\textwidth]{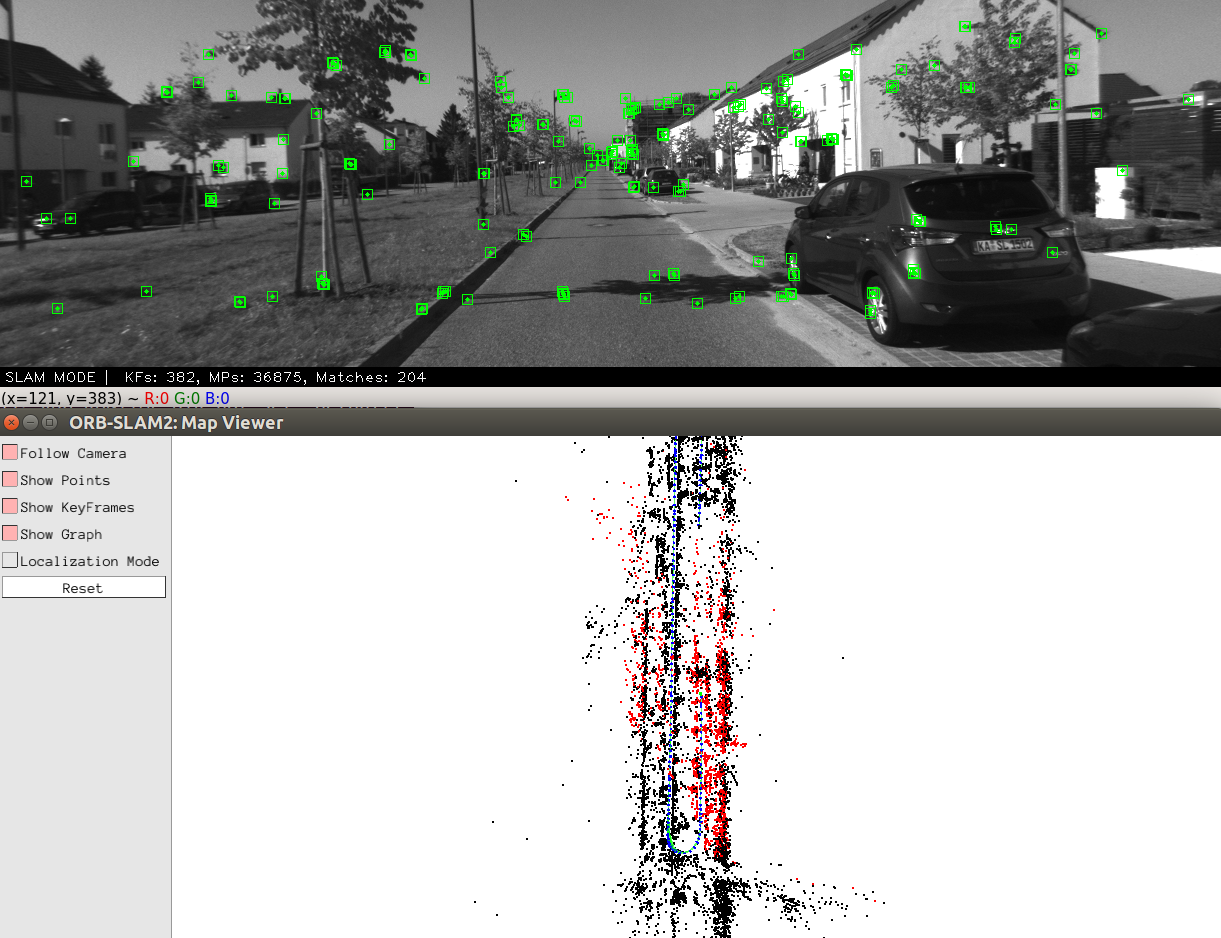}
        \caption{ORB-SLAM2}
    \end{subfigure}
    \begin{subfigure}{.42\textwidth}
        \centering
        \includegraphics[width=0.98\textwidth]{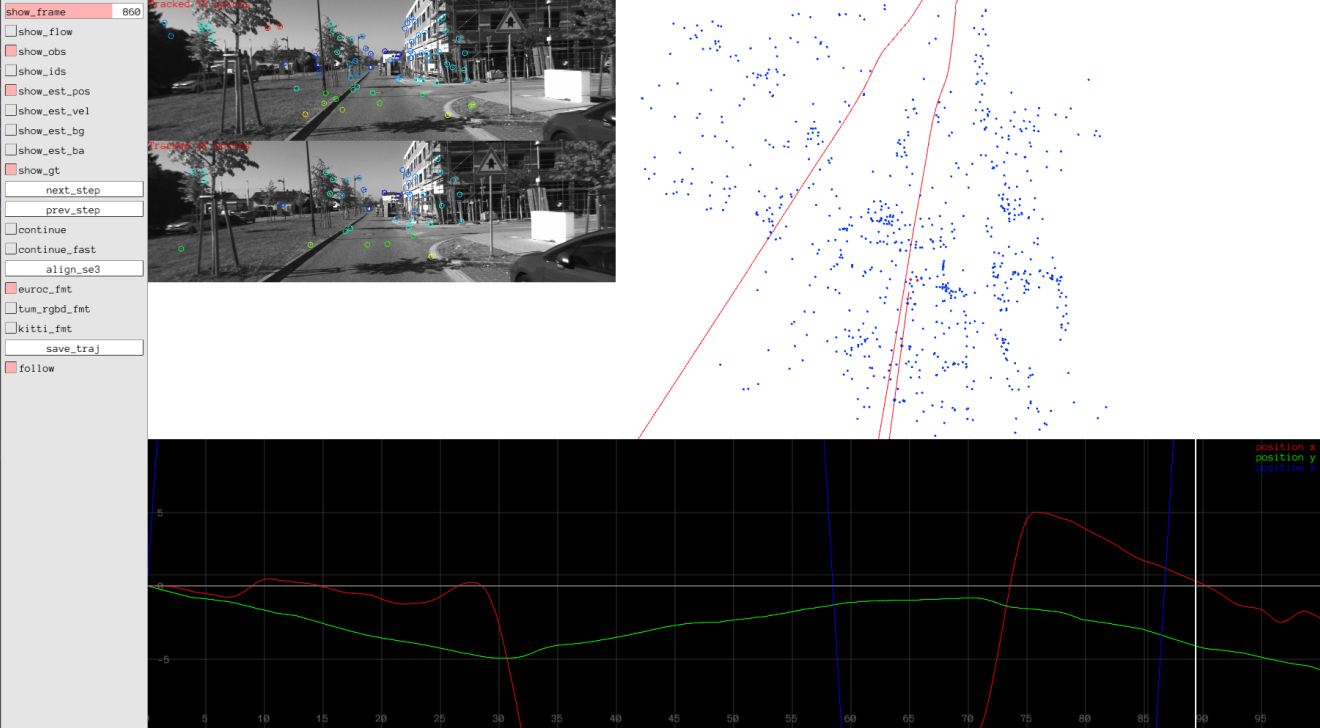}
        \caption{Basalt}
    \end{subfigure}
    \begin{subfigure}{.42\textwidth}
        \centering
        \includegraphics[width=0.98\textwidth]{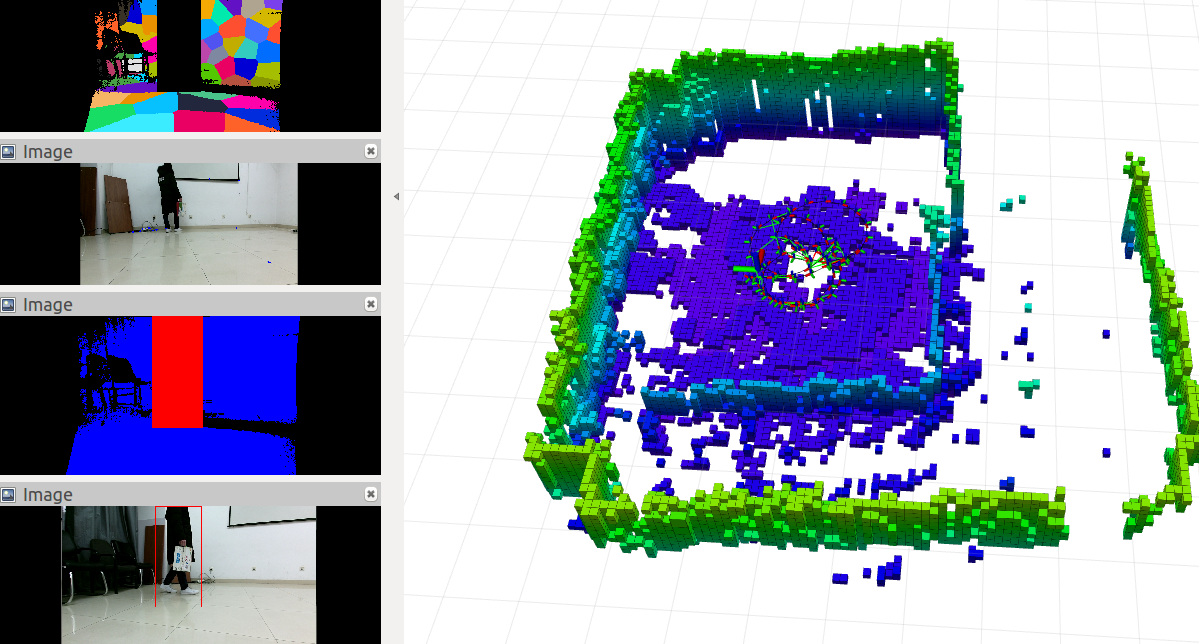}
        \caption{DRE-SLAM}
    \end{subfigure}
    \caption{GUI examples.}
\label{fig_interfaces}
\end{figure}

\subsection{A practical overview of open source solutions}

During the process of launching and working with algorithms, a number of aspects has been identified that must be taken into account when dealing with SLAM solutions.

First, all algorithms were run on Linux-based systems (most often Ubuntu). Launching on Windows seems to be quite time-consuming and, in general, all examples and instructions from the authors are made for Ubuntu or macOS.  Moreover, it was found that a number of algorithms require older versions of Ubuntu (16.04 or 18.04). In some cases, this can be critical for a successful launch.

Secondly, algorithms have different interfaces. The most unified way is to use the Robot Operating System (ROS). In this case, everything depends on the support of the developers of the algorithm. The presence of a separate community, pipeline for data usage, and many examples make ROS a convenient option for launching. Other options are graphical interfaces used by developers (e.g. Pangolin viewer). The quality of the viewer implementation and interface depends on the developer and may differ from one algorithm to another.

Thirdly, the points mentioned above make it especially important to have support from the authors of the SLAM systems. It can be in form of answers to issues on GitHub, availability of documentation and examples of launching. In some cases, the authors have their own website with documentation, or even a channel on Slack to discuss the algorithm and emerging problems \cite{openvslam}.

Fourth, the most convenient way to run SLAM algorithms is to use a ready-made Docker image. Just a few authors provide such a solution. To simplify the launch of algorithms and to the benefit of the community, a publicly available \cite{dockers} repository with docker images has been created and uploaded to the Docker Hub.

\subsection{Datasets description}
\begin{figure}[h]
\centering
    \begin{subfigure}{.42\textwidth}
        \centering
        \includegraphics[width=0.9\textwidth]{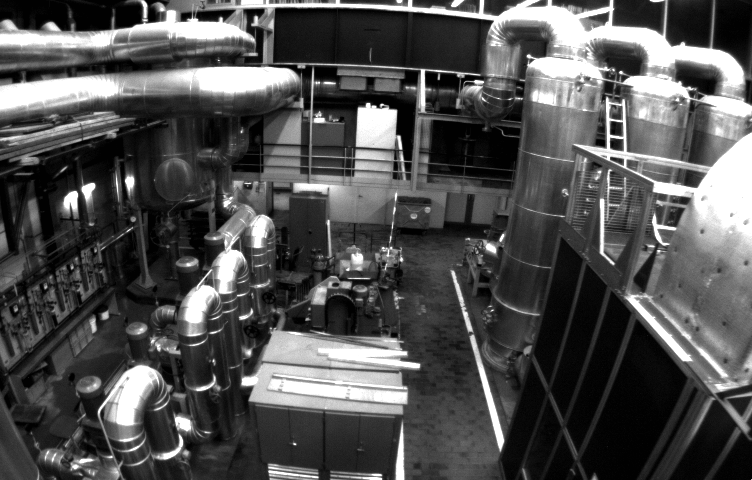}
        \caption{EuRoC MAV}
    \end{subfigure}
    \begin{subfigure}{.47\textwidth}
        \centering
        \includegraphics[width=0.39\textwidth ]{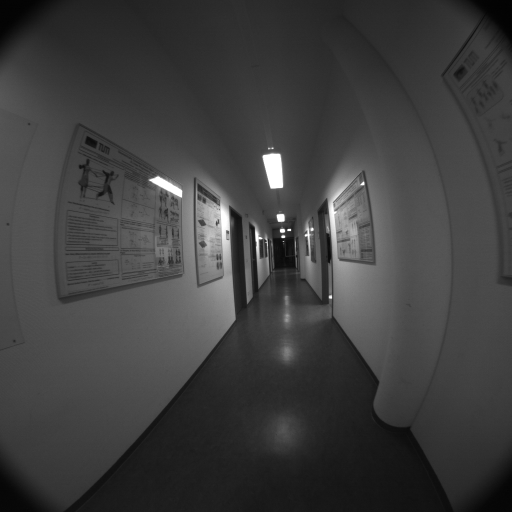}
        \includegraphics[width=0.39\textwidth]{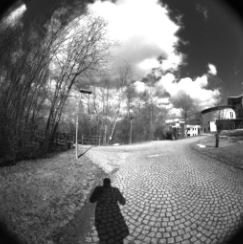}
        \caption{TUM VI}
    \end{subfigure}
    
    \begin{subfigure}{.47\textwidth}
        \centering
        \includegraphics[width=0.8\textwidth]{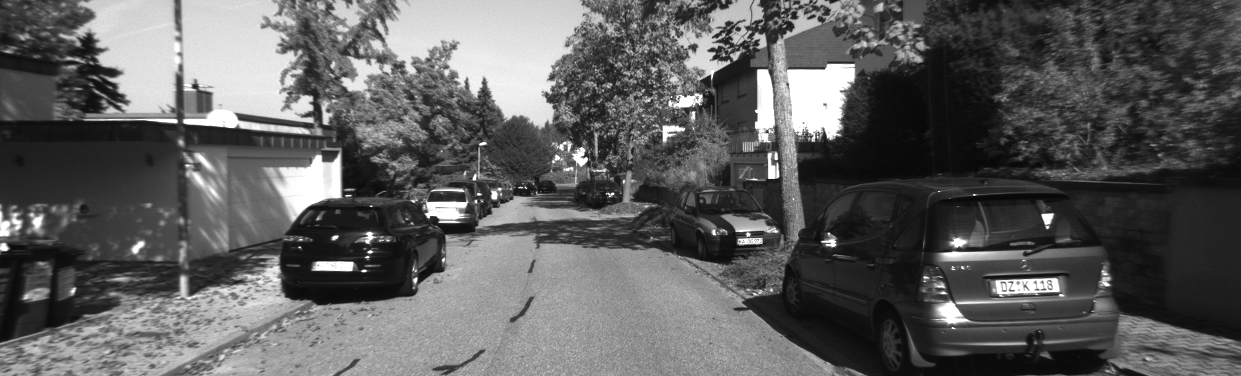}
        \caption{KITTI}
    \end{subfigure}
    
    \begin{subfigure}{.47\textwidth}
        \centering
        \includegraphics[height=.9in ]{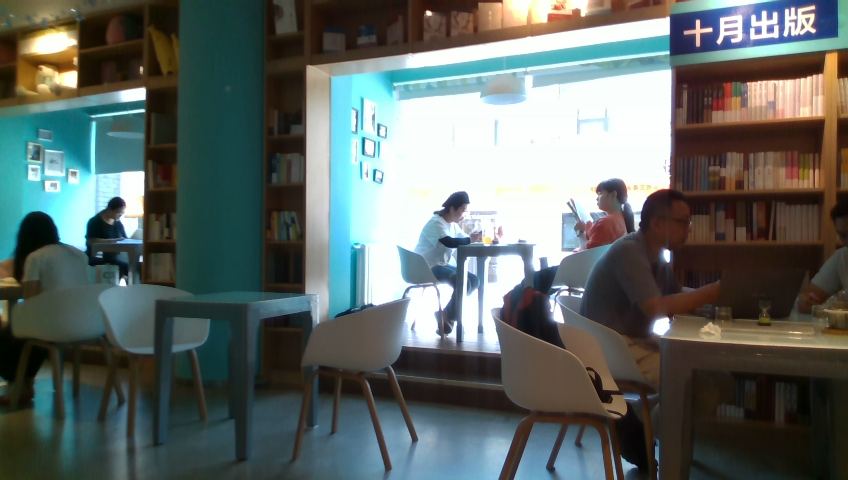}
        \includegraphics[height=.9in]{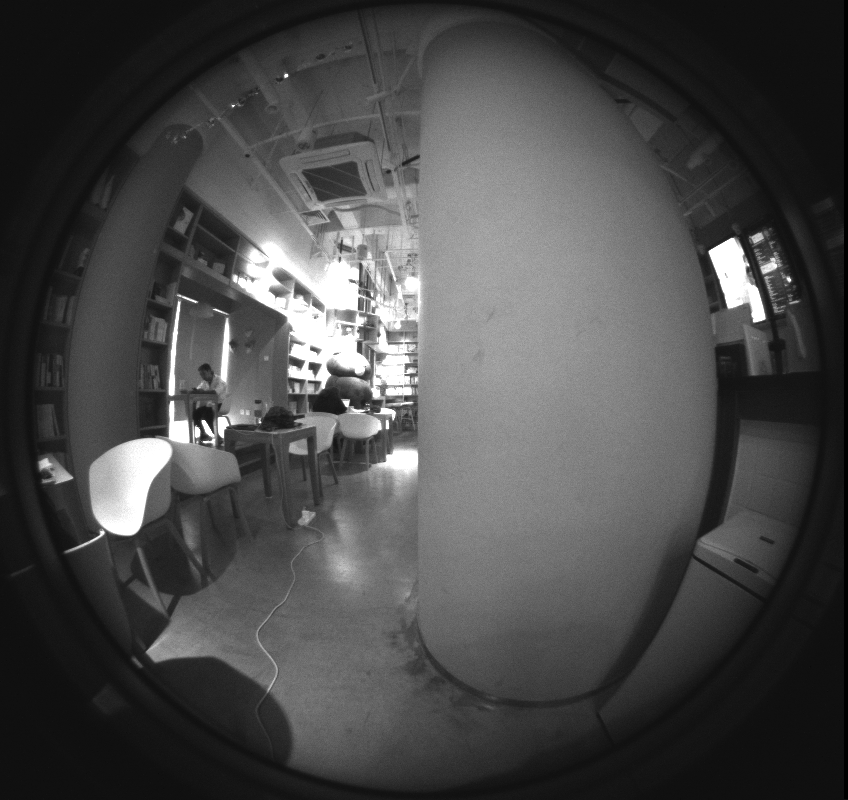}
        \caption{OpenLoris-Scene}
    \end{subfigure}
    
    \begin{subfigure}{.47\textwidth}
        \centering
        \includegraphics[height=0.8in ]{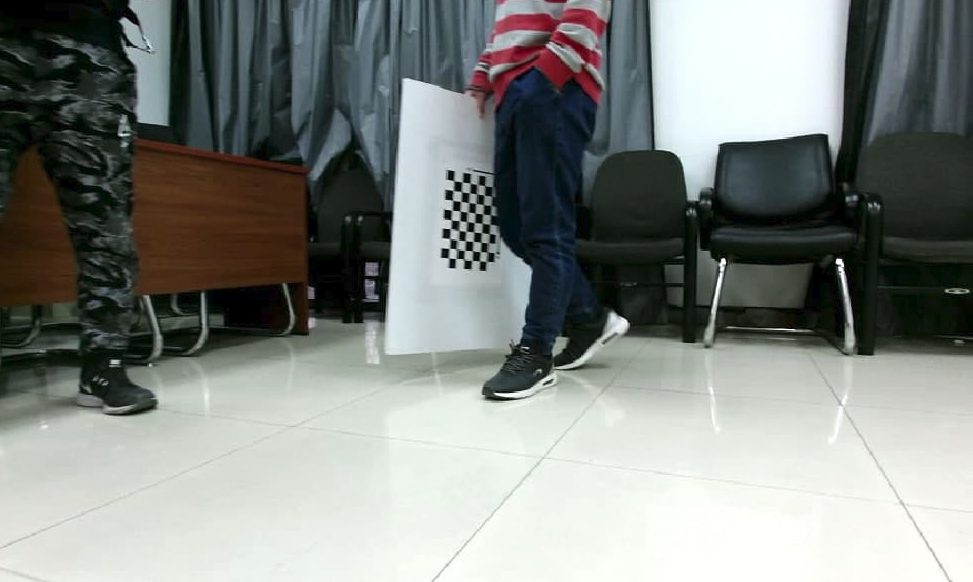}
        \includegraphics[height=0.8in]{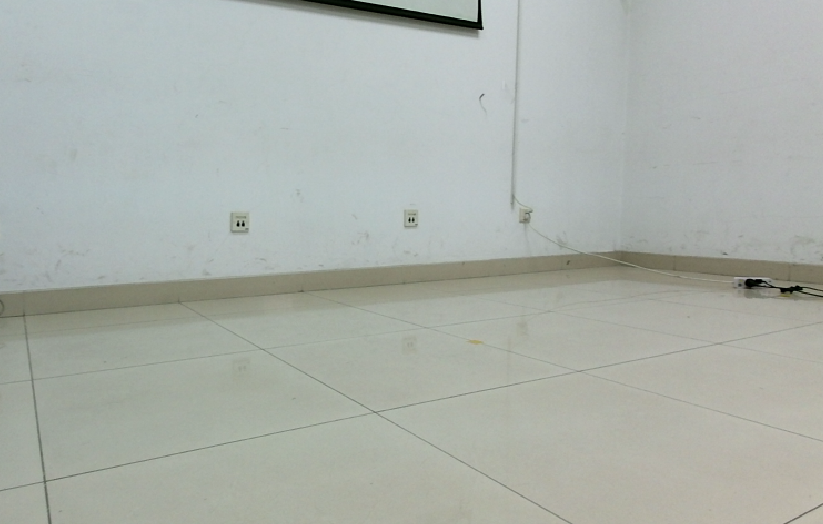}
        \caption{DRE dataset}
    \end{subfigure}
    
    \caption{Examples of images from the datasets.}
\label{fig_darasets}
\end{figure}

Datasets are actively used for SLAM algorithms validation and comparison. They should include ground truth information and have a redundant amount of sensors in order to test different SLAM approaches based on different sensor stacks.

This paper reviews visual and visual-inertial SLAM approaches, and therefore one of the main dataset choice factors was camera image presence. 
Datasets were examined for the existence of loop closures and additional sensors (IMU, GPS, or wheel encoders). This section does a short description of all the datasets that have been considered. The final choice for the experiments is five datasets: EuRoC MAV, TUM VI, KITI, Open Loris, and DRE dataset. Camera data is shown in Fig. \ref{fig_darasets}.

\begin{table*}[h]
\begin{center}
\centering
\caption{Sensor setup for dataset collection.}
\label{table:dataset_sensors}
\resizebox{\textwidth}{!}{%
\begin{tabular}{|c|c|c|c|c|c|c|}
\hline \bfseries Dataset name & \bfseries Camera & \bfseries GPS & \bfseries IMU & \bfseries Lidar & \bfseries Wheel odometry & \bfseries \Centerstack{Additional \\ sensors} \\ 

\hline EuRoC MAV & \Centerstack{Aptina MT9V034 global shutter, \\ WVGA monochrome (2×20 Hz)} & x & ADIS16448 (200 Hz) & x & x & x \\ 

\hline TUM VI & 1 stereo gray (20 Hz) & x & BMI160 (200 Hz) & x & x & x \\

\hline KITTI & \Centerstack{2x Point Grey FL2-14S3C-C \\ global shutter (15 Hz)} & \Centerstack{OXTS RT 3003 \\ (250 Hz)} & OXTS RT 3003 (250 Hz) & \Centerstack{Velodyne HDL-64E \\ (5–20 Hz)} & x & x \\ 

\hline OpenLoris-Scene & \Centerstack{RealSense D435i (30 Hz) \\ RealSense T265 (30 Hz)} & x & RealSense (60-400 Hz) & \Centerstack{Hokuyo UTM30LX \\ (40 Hz)} & Odometer (20 Hz) & x \\ 

\hline DRE dataset & Kinect 2.0 (20 Hz) & x & x & x & \Centerstack{Wheel \\ encoders (100 Hz)} & x \\ 

\hline KAIST & \Centerstack{2x FLIR FL3-U3-20E4C-C \\ global shutter (2x10 Hz)} & \Centerstack{U-Blox \\ EVK-7P (10 Hz)} & Xsens MTi-300 (200 Hz) & \Centerstack{Velodyne VLP-16 (10 Hz), \\ SICK LMS-511 (100 Hz)} & \Centerstack{2x RLS LM13 \\ encoders (100 Hz)} & \Centerstack{VRS–GPS, \\ Three-axis FOG, \\ Altimeter} \\

\hline Malaga Urban & 1 stereo RGB (20 Hz) & DELUO (1 Hz) & xSens MTi (100 Hz) & \Centerstack{3x Hokuyo UTM-30LX (25 Hz), \\ 2x SICK LMS-200 (75 Hz)} & x & x \\ 

\hline Oxford RobotCar & \Centerstack{BBX3-13S2C-38 \\ global shutter (16 Hz), \\ 3x GS2-FW-14S5C-C \\ global shutter (11.1 Hz)} & \Centerstack{NovAtel \\ SPAN-CPT (50 Hz)} & \Centerstack{NovAtel \\ SPAN-CPT (50 Hz)} & \Centerstack{2 x SICK LMS-151 (50 Hz) \\ 1 x SICK LD-MRS (12.5 Hz)} & x & x \\ 

\hline RPNG OpenVINS & Aptina MT9V034 (60 Hz) & x & ADIS16448 (200 Hz) & x & x & x \\ 

\hline Segway DRIVE & RealSence ZR300 (30 Hz) & x & BMI055 (250 Hz) & \Centerstack{Hokuyo UTM30LX \\ (40 Hz)} & x & x \\ 

\hline TUM RGB-D & Kinect (30 Hz) & x & x & x & x & x \\ 

\hline UMich NCLT & 6 RGB (omni) (5 Hz) & Garmin 18x (5 Hz) & \Centerstack{Microstrain \\ 3DM-GX3-45 (100 Hz)}  & \Centerstack{Velodyne HDL-32E (10 Hz) \\ Hokuyo UTM-30LX (40 Hz)\\ Hokuyo URG-04LX (10 Hz)} & Wheel encoders & \Centerstack{FOG, \\ RTK GPS} \\ 

\hline Zurich Urban MAV & RGB rolling shutter (30 Hz) & GPS & IMU (10 Hz) & x & x & x \\ \hline 

\end{tabular}}
\end{center}
\end{table*}

\begin{table*}[h]
\begin{center}
\centering
\caption{Dataset main features.}
\label{table:feats1}
\resizebox{\textwidth}{!}{%
\begin{tabular}{|c|c|c|c|c|}
\hline \bfseries Dataset name & \bfseries Year & \bfseries ROS bag & \bfseries Ground truth & \bfseries Indoor / Outdoor \\ 

\hline EuRoC MAV & 2016 & + & Motion capture system (acc. $\approx$ 1 mm) & Indoor \\ 

\hline TUM VI & 2018 & + & Motion capture system (acc. $\approx$ 1 mm) & Indoor / Outdoor \\

\hline KITTI & 2013 & + (parser is provided) & OXTS RT 3003 inertial navigation system (INS) (acc. $<$ 10 cm) & Outdoor \\

\hline OpenLoris-Scene & 2019 & + & OptiTrack motion capture system / LIDAR SLAM (acc. $<$ 10 cm) & Indoor \\ 

\hline DRE dataset & 2019 & + & Downview camera and markers (acc. unknown) & Indoor\\ 

\hline KAIST & 2019 & + (rosbag player) & LIDAR SLAM (acc. unknown) & Outdoor \\ 

\hline Malaga Urban & 2014 & - & GPS (low acc.) & Outdoor \\

\hline Oxford RobotCar & 2017 & - & SPAN GNSS Inertial Navigation System (acc. $<$ 10 cm) & Outdoor \\ 

\hline RPNG OpenVINS & 2019 & + & GPS in outdoor scenes (low acc.) & Indoor / Outdoor \\ 

\hline Segway DRIVE & 2019 & + & LIDAR SLAM (acc. $<$ 10 cm) & Indoor \\ 

\hline TUM RGB-D & 2012 & + & Motion capture system (acc. $\approx$ 1 mm) & Indoor \\ 

\hline UMich NCLT & 2015 & + (parser is provided) & Fused GPS/IMU/LIDAR (acc. $\approx$ 10 cm) & Outdoor \\ 

\hline Zurich Urban MAV & 2017 & + (parser is provided) & Pix4D SLAM (acc. unknown) & Outdoor \\ \hline

\end{tabular}}
\end{center}
\end{table*}

\begin{table}[h]
\begin{center}
\centering
\caption{Datasets` additional features.}
\label{table:feats2}
\setlength\tabcolsep{5pt}
\resizebox{0.48\textwidth}{!}{
\begin{tabular}{|c|c|c|c|c|c|}
\hline \bfseries Dataset name & \bfseries \Centerstack{Variable weather \\ conditions} & \bfseries \Centerstack{Dynamic \\ objects} & 
\bfseries \Centerstack{Memory \\ consumption} & \bfseries \Centerstack{Path \\ length} \\ 

\hline EuRoC MAV & - & - & 1-2.5 Gb & 30-130 m \\ 

\hline TUM VI & - & + & 3-60 Gb & $<$20 km \\ 

\hline KITTI & + & + & $>$20 Gb & $<$40 km \\ 

\hline OpenLoris-Scene & - & + & 6-33 Gb & Not mentioned \\ 

\hline DRE dataset & - & \Centerstack{Static, Low/High Dynamic} & 4-8 Gb & Not mentioned \\ 

\hline KAIST & - & + & 1-30 Gb & 1-30 km \\ 

\hline Malaga Urban & - & + & 1-33 Gb & $<$36.8 km \\ 

\hline Oxford RobotCar & + & + & 10-500 Gb & Not mentioned \\ 

\hline RPNG OpenVINS & - & - & 1-2.6 Gb & 27-105 m and 2.3, 7.4 km \\ 

\hline Segway DRIVE & - & + & 1-20 Gb & 50 km \\ 

\hline TUM RGB-D & - & + & 0.2-2.5 Gb & 0.4 km \\ 

\hline UMich NCLT & + & + & 80-110 Gb & 1-7.5 km \\ 

\hline Zurich Urban MAV & - & + & 28 Gb & 2 km \\ \hline

\end{tabular}}
\end{center}
\end{table}

A commonly used option for evaluating visual and visual-inertial SLAM algorithms is the EuRoC MAV dataset \cite{EUROC}, but its image resolution and bit depth are not quite state-of-the-art anymore. Also, this dataset is mostly suitable for Micro Air Vehicles and does not include odometry readings suitable for some SLAM approaches for mobile wheeled robots. SLAM article authors usually test and compare their algorithms on the EuRoC dataset primarily but in many cases SLAM algorithms show good performance on this dataset and bad on other datasets. That will be shown later.

TUM VI \cite{TUM} is another popular dataset for evaluating visual and visual-inertial SLAM algorithms. It includes more varied scenes including indoor and outdoor environments and longer sequences than the EuRoC dataset. Dataset provides time-synchronized camera images with 1024x1024 resolution and IMU measurements. However, this dataset provides accurate pose ground truth from a motion capture system only at the start and end of the sequences. There is no ground truth data for the outdoor part of the sequence. 

OpenLoris-Scene dataset \cite{shi2019openlorisscene} is a dataset that aims to help evaluate the maturity of SLAM and scene understanding algorithms for real-world deployment, by providing visual, inertial and odometry data recorded with real robots in real scenes, and ground-truth robot trajectories acquired by motion capture system or high-resolution LiDARs. The important peculiarity of this dataset is quite hard real-life conditions with recordings of the same scenes in different lighting levels and with dynamic content. 

TUM RGB-D SLAM Dataset \cite{tum_rgbd} is a dataset that includes RGB-D data and ground-truth data with the goal to establish a novel benchmark for the evaluation of visual odometry and visual SLAM systems. The dataset contains the color and depth images of a Microsoft Kinect sensor and accelerometer data. The ground-truth trajectory was obtained from a high-accuracy motion-capture system.

KITTI \cite{KITTI} vision dataset was collected using an autonomous driving platform and scenes are captured by driving around the mid-size city of Karlsruhe, in rural areas and on highways. Dataset has a benchmark suite that includes odometry, object detection and tracking, road \cite{kitti_road}, stereo and flow benchmarks. The odometry dataset was chosen for visual SLAM evaluation. This benchmark consists of 22 stereo sequences, saved in loss-less {\em png} format and has a leaderboard. For this benchmark, it is possible to provide results using monocular or stereo visual odometry, laser-based SLAM or algorithms that combine visual and LIDAR information. The data was recorded using an eight-core i7 computer.

KAIST Urban Data Set \cite{KAIST} is another dataset collected using a car platform. The vehicle was equipped with two 2D and two 3D LiDARs to collect data on the surrounding environment. Additionally, the sensor suite included a stereo camera installed facing the front of the vehicle, GPS, VRS GPS, IMU, Fiber Optic Gyro (FOG) and altimeter. All sensor information was provided in a raw file format with timestamps. Three PCs were used to collect the data. The system clocks of the three PCs were periodically synchronized using the Chrony library \cite{chrony}. Each PC used an i7 processor, 512 GB solid-state drive (SSD), and 32 GB DDR4 memory.

Oxford RobotCar Dataset \cite{oxford_car} contains over 100 repetitions of a consistent route through Oxford, UK, captured over a period of over a year. The dataset captures many different combinations of weather, traffic and pedestrians, along with longer term changes such as construction and roadworks. 

RPNG OpenVINS Dataset was developed by OpenVINS \cite{openvins} authors. It includes ArUco datasets built using Synchronized Visual-Inertial Sensor System \cite{device_openvins} in an indoor environment with ArUco markers. Also, this dataset includes two long sequences (2.3 and 7.4 km) build using ironsides \cite{PIRVS} visual-inertial sensor.

DRE dataset \cite{dre} is a dataset collected using the Redbot robot. It was made by DRE SLAM \cite{dre} authors for their algorithm evaluation. One of the key features of this dataset is the presence of odometry readings from wheel encoders. Moreover, the dataset includes scenes with low and high dynamic objects.

Segway DRIVE benchmark \cite{seg_dataset} includes datasets collected by Segway delivery robots deployed in real office buildings and shopping malls. Each robot was equipped with a global-shutter fisheye camera, a consumer-grade IMU synced to the camera on chip, two low-cost wheel encoders, and a removable high-precision lidar.

The Zurich Urban Micro Aerial Vehicle Dataset \cite{Zurich_dataset} was collected using camera equipped Micro Aerial Vehicle (MAV) flying within urban streets at low altitudes (i.e., 5-15 meters above the ground). The 2 km dataset consists of time-synchronized aerial high-resolution images, GPS and IMU sensor data, ground-level street view images, and ground truth data.

The Malaga Stereo and Laser Urban Dataset \cite{Malaga_dataset} was gathered entirely in urban scenarios with a car equipped with several sensors, including one stereo camera (Bumblebee2) and five laser scanners. One distinctive feature of this dataset is high-resolution stereo images.

The University of Michigan North Campus Long-Term Vision and LIDAR Dataset \cite{Michigan_dataset} consists of omnidirectional imagery, 3D lidar, planar lidar, GPS, and proprioceptive sensors for odometry collected using a Segway robot.

The following Table \ref{table:dataset_sensors} shows sensors used in the described earlier datasets. Datasets main and additional features are shown in the Tables \ref{table:feats1} and \ref{table:feats2}. As it could be seen from the tables, there is no perfect dataset that matches different demands, such as indoor/outdoor data, changing conditions, dynamic content, several types of sensors, etc. As well as there is no unified way of data organisation. Several standards of ground truth formats exist (TUM, KITTY, EuRoC), but only this. It means that there is always a need to use several datasets and tune data formats for the algorithms which are itself demands special conditions for data.



\subsection{Metrics}

The result of the SLAM system is a trajectory and a map of the surrounding area. Despite the fact that it is theoretically possible to compare the resulting maps, in practice, obtaining reference maps is not a trivial task. In addition, different systems use different map formats, making this task even more difficult. For this reason, most often comparisons are made using the resulting trajectories.

However, it should be noted that a good trajectory does not necessarily mean a good map. For example, a map might be sparse with a low number of features or landmarks whereas a trajectory is precise. Small errors on the map, which have little effect on localization accuracy, can interfere with the normal functioning of the robot, for example, by blocking movement in a doorway or corridor.

Readings from additional sensors can be used to obtain reference trajectories. For outdoor datasets, high-precision GNSS systems are used in conjunction with the IMU sensor \cite{geiger2013vision}, while indoor optical tracking systems \cite{sturm2012benchmark} can be used.

Relative Positional Error (RPE) \cite{Kummerle2011} measures the trajectory drift over a fixed time interval $ \Delta $ in a reference frame, thus it does not accumulate previous errors. Additionally, many researchers tend to separate translational and rotational errors because it gives a more clear understanding of SLAM algorithm quality and it is not plausible to mix angles with poses (degrees with meters). For pose predictions of SLAM algorithm $ \mathbf {P_1}, ... , \mathbf{P_n} \in \mathrm {SE} (3)$, and the ground-true poses $ \mathbf {Q_1}, ... , \mathbf{Q_n} \in \mathrm {SE} (3) $ the relative error for each point in time interval is defined as in \cite{Benchmarking_Comparing}:

\begin{equation}
    \mathbf{E}_i = \left( \mathbf{Q}^{-1}_i \mathbf{Q}_{i + \Delta} \right)^{-1} \left( \mathbf{P}^{-1}_i \mathbf{P}_{i + \Delta} \right)^{-1}
\end{equation} 

Thus, for a sequence of $ n $ measurements and their corresponding positions, there are $ m = n - \Delta $ relative errors. Using the error data, the root mean square error is calculated:

\begin{equation}
\mathrm{RMS}\left(\mathbf{E}_{1:n}, \Delta \right) = \left( \frac{1}{m} \sum^m_{i = 1} 
\lVert{\mathbf{E}_i}\rVert^2  \right)^{1/2}
\end{equation}

In practice, it is useful to know Relative Rotational and Relative Translation errors for time interval $ \Delta $. It allows estimating the quality of the SLAM algorithm for a particular situation without the influence of previous movements. Thus, rotational and translational parts can be calculated separately:

\begin{equation}
\mathrm{RMS}\left(\mathbf{E}^{trans}_{1:n}, \Delta \right) = \left( \frac{1}{m} \sum^m_{i = 1} 
\lVert{\mathbf{E}^{trans}_i}\rVert^2  \right)^{1/2}
\end{equation}

\begin{equation}
\mathrm{RMS}\left(\mathbf{E}^{rot}_{1:n}, \Delta \right) = \left( \frac{1}{m} \sum^m_{i = 1} 
\lVert{\mathbf{E}^{rot}_i}\rVert^2  \right)^{1/2}
\end{equation}
where $ \mathbf{E}^{trans}_i$ refers to the translation components of the Relative Pose Error and ${\mathbf{E}^{rot}_i}$ refers to the rotational part.



In some cases, instead of the mean square, the median or means of other orders can be used. For example, the arithmetic mean will be less sensitive to anomalies than the root mean square, but more sensitive than the median.

For SLAM the global consistency of the resulting trajectory is also important, which can be estimated by calculating the difference between the corresponding estimated and reference poses (Absolute error) in a global frame. Since the frames of reference for the estimated and reference trajectories may differ from each other, the first step is to align them with each other. This can be done using the analytical solution provided in \cite{horn1987closed}. The solution finds a transformation $ \mathbf {S} \in \mathrm {SE} (3) $ that minimizes the root-mean-square difference between $ \mathbf {P} _ {1: n} $ and $ \mathbf {Q} _ {1: n } $. Using this transformation, the difference between the poses can be calculated as:
\begin{equation}
\begin{gathered}
\mathbf{F}_i = \mathbf{Q}^{-1}_i \mathbf{S} \mathbf{P}_i \\
\mathrm{RMS}\left(\mathbf{F}_{1:n} \right) = \left(\frac{1}{m} \sum^m_{i = 1} \lVert{\mathbf{F}_i}\rVert^2  \right)^{1/2}
\end{gathered}
\end{equation}

In practice, both metrics show a strong correlation with each other. But the main interest is in the comparison of different trajectories with one reference ground-truth trajectory for the whole length. Therefore, the more intuitive Absolute Error is used most often as well as in the results section.
For the calculation of trajectories and metrics, the comparison EVO tool \cite{evo} was used. Exhaustive documentation, visualisation possibilities, and easy-to-use interface make it very convenient to compare a large number of results. Another possible way to track the performance is to use the KITTI Vision Benchmark Suite \cite{KITTI} web page. It allows seeing the translation and rotation error of an algorithm in a web table with more than a hundred different solutions, but it is limited to one dataset.

\section{Results}



\begin{table}[h]
\begin{center}
\centering
\caption{Results of SLAM algorithms tested on EUROC dataset}
\label{table:euroc_table}
\begin{adjustbox}{width=0.48\textwidth}
\begin{tabular}{|c|c|c|c|c|c|c|c|c|} 
\hline
\textbf{}            & \multicolumn{4}{c|}{\textbf{V1\_01 }}                                                                                                                                                                                                         & \multicolumn{4}{c|}{\textbf{MH\_05}}                                                                                                                                                                             \\ 
\cline{2-9}
\textbf{Framework}   & \multicolumn{2}{c|}{\textbf{position(m) }}                                                                                           & \multicolumn{2}{c|}{\textbf{rotation(deg) }}                                                           & \multicolumn{2}{c|}{\textbf{position(m)}}                                                              & \multicolumn{2}{c|}{\textbf{rotation(deg)}}                                                             \\ 
\cline{2-9}
\textbf{}            & \textbf{max}                                                                & \begin{tabular}[c]{@{}c@{}}\textbf{rms}\\\end{tabular} & \textbf{max}                                  & \begin{tabular}[c]{@{}c@{}}\textbf{rms}\\\end{tabular} & \textbf{max}                                  & \begin{tabular}[c]{@{}c@{}}\textbf{rms}\\\end{tabular} & \textbf{max}                                  & \begin{tabular}[c]{@{}c@{}}\textbf{rms}\\\end{tabular}  \\ 
\hline
\textbf{ORB-SLAM2}   & 0.16                                                                        & 0.08                                                   & 13.97                                         & \textbf{\textit{\textcolor[rgb]{0.502,0,0.502}{4.70}}}                                                   & 0.15                                          & 0.05                                                   & 20.93                                         & 6.60                                                    \\ 
\hline
\textbf{MapLab}      & 0.30                                                                        & 0.15                                                   & 10.03                                         & 6.13                                                   & 1.13                                          & 0.59                                                   & 2.99                                          & 1.27                                                    \\ 
\hline
\textbf{LDSO}        & 0.16                                                                        & 0.08                                                   & 10.34                                         & \textcolor[rgb]{0,0.502,0}{\textbf{4.68}}                                                   & 1.49                                          & 0.08                                                   & 28.35                                         & 20.45                                                   \\ 
\hline
\textbf{VINS-Mono}   & 0.21                                                                        & 0.09                                                   & 6.77                                          & 6.09                                                   & 0.48                                          & 0.30                                                   & 2.58                                          & \textcolor[rgb]{0.502,0,0.502}{\textbf{\textit{0.79}}}                                                    \\ 
\hline
\textbf{VINS-Fusion} & 0.24                                                                        & 0.07                                                   & 10.19                                         & 5.60                                                   & 0.40                                          & 0.19                                                   & 4.81                                          & 1.87                                                    \\ 
\hline
\textbf{OpenVSLAM}   & 0.15                                                                        & 0.08                                                   & 14.13                                         & 4.73                                                   & \textbf{\textit{\textcolor[rgb]{0.502,0,0.502}{0.14}}}    & \textcolor[rgb]{0,0.502,0}{\textbf{0.04}}              & 20.85                                         & 6.54                                                    \\ 
\hline
\textbf{Basalt}      & \textcolor[rgb]{0,0.502,0}{\textbf{0.05}}                               & \textcolor[rgb]{0,0.502,0}{\textbf{0.03}}              & \textbf{\textcolor[rgb]{0,0.502,0}{5.99}} & 5.36          & 0.17                                          & 0.09                                                   & \textcolor[rgb]{0,0.502,0}{\textbf{1.42}} & \textcolor[rgb]{0,0.502,0}{\textbf{0.67}}           \\ 
\hline
\textbf{Kimera}      & 0.08                                                                        & 0.04                                                   & 6.34                                          & 5.59                                                   & 0.58                                          & 0.34                                                   & 3.82                                          & 2.34                                                    \\ 
\hline
\textbf{OpenVINS}    & 0.11                                                                        & 0.05                                                   & \textcolor[rgb]{0.502,0,0.502}{\textbf{\textit{6.05}}}     & 5.45              & 0.48                                          & 0.16                                                   & 3.36                                          & 1.30                                                    \\ 
\hline
\textbf{ORB-SLAM3}   & \textbf{\textit{ \textcolor[rgb]{0.502,0,0.502}{0.07}} } & \textbf{\textcolor[rgb]{0,0.502,0}{0.03}}          & 6.39                                          & 5.94                                                   & \textbf{\textcolor[rgb]{0,0.502,0}{0.11}} & \textbf{\textcolor[rgb]{0,0.502,0}{0.04}}          & \textcolor[rgb]{0.502,0,0.502}{\textbf{\textit{1.68}}}     & 0.93               \\
\hline
\end{tabular}
\end{adjustbox}
\end{center}
\end{table}

\begin{table}[h]
\begin{center}
\centering
\caption{Results of SLAM algorithms tested on DRE dataset.}
\label{table:dre}
\begin{adjustbox}{width=0.48\textwidth}
\begin{tabular}{|c|c|c|c|c|c|c|c|c|} 
\hline
\textbf{}                                                                   & \multicolumn{4}{c|}{\begin{tabular}[c]{@{}c@{}}\textbf{Low Dynamic (ST2)}\\\end{tabular}}                                                                                     & \multicolumn{4}{c|}{\textbf{High Dynamic (HD2)}}                                                                                                                                \\ 
\cline{2-9}
\textbf{Framework}                                                          & \multicolumn{2}{c|}{\textbf{position }}                                               & \multicolumn{2}{c|}{\textbf{rotation }}                                               & \multicolumn{2}{c|}{\textbf{position }}                                               & \multicolumn{2}{c|}{\textbf{rotation }}                                                 \\ 
\cline{2-9}
                                                                            & \textbf{max}                              & \textbf{rms}                              & \textbf{max}                              & \textbf{rms}                              & \textbf{max}                              & \textbf{rms}                              & \textbf{max}                               & \textbf{rms}                               \\ 
\hline
\textbf{ORB-SLAM2 mono}                                                     & 0.27                                      & 0.09                                      & 3.95                                      & 1.62                                      & 1.12                                      & 0.31                                      & 39.77                                      & 16.49                                      \\ 
\hline
\begin{tabular}[c]{@{}c@{}}\textbf{\textbf{ORB-SLAM2 rgb-d}}\\\end{tabular} & 0.16                                      & 0.09                                      & 5.14                                      & 2.05                                      & 1.23                                      & 0.70                                      & 45.55                                      & 19.30                                      \\ 
\hline
\begin{tabular}[c]{@{}c@{}}\textbf{DRE-SLAM}\\\end{tabular}                 & \textcolor[rgb]{0,0.502,0}{\textbf{0.05}} & \textcolor[rgb]{0,0.502,0}{\textbf{0.01}} & \textcolor[rgb]{0,0.502,0}{\textbf{2.51}} & \textcolor[rgb]{0,0.502,0}{\textbf{0.85}} & \textcolor[rgb]{0,0.502,0}{\textbf{0.26}} & \textcolor[rgb]{0,0.502,0}{\textbf{0.06}} & \textcolor[rgb]{0,0.502,0}{\textbf{19.85}} & \textcolor[rgb]{0,0.502,0}{\textbf{4.94}}  \\
\hline
\end{tabular}
\end{adjustbox}
\end{center}
\end{table}

\begin{table*}[h]
\begin{center}
\centering
\caption{Results of SLAM algorithms tested on TUM VI dataset.}
\label{table:TUM_VI_table}
\resizebox{\textwidth}{!}{
\begin{tabular}{|c|c|c|c|c|c|c|c|c|c|c|c|c|} 
\hline
                   & \multicolumn{4}{c|}{\textbf{Corridor 1}}                                                                                                                                                      & \multicolumn{4}{c|}{\textbf{Magistrale 1}}                                                                                                                                                    & \multicolumn{4}{c|}{\textbf{Room 1}}                                                                                                                                                           \\ 
\cline{2-13}
\textbf{Framework} & \multicolumn{2}{c|}{\textbf{position(m)}}                                                     & \multicolumn{2}{c|}{\textbf{rotation(deg)}}                                                   & \multicolumn{2}{c|}{\textbf{position(m)}}                                                     & \multicolumn{2}{c|}{\textbf{rotation(deg)}}                                                   & \multicolumn{2}{c|}{\textbf{position(m)}}                                                     & \multicolumn{2}{c|}{\textbf{rotation(deg)}}                                                    \\ 
\cline{2-13}
                   & \textbf{max}                                  & \textbf{rms}                                  & \textbf{max}                                  & \textbf{rms}                                  & \textbf{max}                                  & \textbf{rms}                                  & \textbf{max}                                  & \textbf{rms}                                  & \textbf{max}                                  & \textbf{rms}                                  & \textbf{max}                                  & \textbf{rms}                                   \\ 
\hline
\textbf{MapLab}    & \textcolor[rgb]{0.502,0,0.502}{\textbf{\textit{0.78}}}     & 0.27     & 4.18                                          & 2.85                                          & 2.72                                          & 0.88                                          & 13.92                                         & 6.37                                          & 0.27                                          & 0.14                                          & 6.05                                          & 2.10                                           \\ 
\hline
\textbf{VINS-Mono} & 1.74                                          & 0.44                                          & \textcolor[rgb]{0,0.502,0}{\textbf{1.40}} & \textcolor[rgb]{0.502,0,0.502}{\textbf{\textit{0.47}}}     & \textcolor[rgb]{0,0.502,0}{\textbf{0.35}} & \textcolor[rgb]{0,0.502,0}{\textbf{0.24}} & \textcolor[rgb]{0,0.502,0}{\textbf{3.64}} & \textcolor[rgb]{0,0.502,0}{\textbf{1.06}} & \textcolor[rgb]{0.502,0,0.502}{\textbf{\textit{0.12}}}     & \textcolor[rgb]{0.502,0,0.502}{\textbf{\textit{0.05}}}     & \textcolor[rgb]{0,0.502,0}{\textbf{3.79}} & 1.74                                           \\ 
\hline
\textbf{OpenVSLAM} & 0.91                                          & \textcolor[rgb]{0.502,0,0.502}{\textbf{\textit{0.23}}}                                          & 135.99                                        & 43.61                                         & 1.26                                          & 0.51                                          & 154.69                                        & 41.44                                         & 0.13                                          & 0.07                                          & 102.54                                        & 32.72                                          \\ 
\hline
\textbf{Basalt}    & 1.34                                          & 0.33                                          & 4.25                                          & 2.32                                          & 1.91                                          & 0.68                                          & \textcolor[rgb]{0.502,0,0.502}{\textbf{\textit{5.28}}}     & \textcolor[rgb]{0.502,0,0.502}{\textbf{\textit{1.47}}}     & 0.23                                          & 0.09                                          & 5.92                                          & \textcolor[rgb]{0.502,0,0.502}{\textbf{\textit{1.25}}}      \\ 
\hline
\textbf{OpenVINS}  & 1.22                                          & 0.36                                          & 5.62                                          & 3.43                                          & 2.68                                          & 0.88                                          & 30.25                                         & 20.02                                         & 0.20                                          & 0.07                                          & 9.69                                          & 5.25                                           \\ 
\hline
\textbf{ORB-SLAM3} & \textbf{\textcolor[rgb]{0,0.502,0}{0.02}} & \textcolor[rgb]{0,0.502,0}{\textbf{0.01}} & \textcolor[rgb]{0.502,0,0.502}{\textbf{\textit{1.48}}}     & \textbf{\textcolor[rgb]{0,0.502,0}{0.45}} & \textcolor[rgb]{0.502,0,0.502}{\textbf{\textit{0.57}}}     & \textcolor[rgb]{0.502,0,0.502}{\textbf{\textit{0.33}}}     & 6.66                                          & 3.69                                          & \textbf{\textcolor[rgb]{0,0.502,0}{0.02}} & \textcolor[rgb]{0,0.502,0}{\textbf{0.01}} & \textcolor[rgb]{0.502,0,0.502}{\textbf{\textit{5.74}}}     & \textcolor[rgb]{0,0.502,0}{\textbf{0.42}}  \\
\hline
\end{tabular}
}
\end{center}
\end{table*}

The tables \ref{table:euroc_table},  \ref{table:dre}, \ref{table:TUM_VI_table}, \ref{table:kitti_table}, \ref{table:open_loris_table} represent the Absolute Pose Error metric (maximum and root mean squared values) for the considered algorithms. The metrics is both for position error in meters and for rotation error in degrees. The lowest error result among all algorithms is marked with bold green numbers and italic purple numbers as the second result. The first three tables represent results on popular and very common datasets in the robotics community. Not all algorithms are present in all tables. Absence means the inability to run on certain types of data, problems with parameters or algorithms themselves. Experiments on Open LORIS data and DRE dataset have been carried out for checking results in harsh conditions (sharp turns, textureless data, dynamic content). If a SLAM system could not be initialized or loses a solution it is marked as "X". The algorithms in the tables are placed in the order of their release date. The examples of obtained trajectories of algorithms on EuRoC MAV mh5 and TUM VI corridor 1 sequence are illustrated in Fig. \ref{fig_trajectories}.

Table \ref{table:euroc_table} illustrates Basalt to be a winner on EuRoC V1\textunderscore01{\textunderscore}easy sequence, and ORB-SLAM3 and OpenVSLAM on MH\textunderscore05{\textunderscore}difficult. Worth reminding, "easy" EuRoC sequences differs from "difficult" one by jiggles and light conditions because of quadcopter data recording. ORB-SLAM3 and OpenVSLAM showed better results in worse conditions. In general, most of the algorithms show low errors and comparable results. Also, a much bigger error of LDSO can be noticed - the only Dense SLAM method which has been used.  

Table \ref{table:TUM_VI_table} represents errors on TUM VI data sequence. Unfortunately, not all methods are capable to work with fish-eye data. The leader here is ORB-SLAM3 with errors, an order of magnitude less than others. It is worth noting, that this approach is 2 years newer than MapLab and VINS Mono - the second leaders, whereas all other algorithms in the table have similar release time. VINS Mono has two times higher error than MapLab on corridor\textunderscore1 sequence whereas it significantly wins on room1 and magistrale\textunderscore1 sequences. The length of magistrale\textunderscore1 sequence is much bigger (around 1 km) than others which makes it especially difficult for the algorithms. Even though OpenVSLAM shows good results on other sequences, here it is used to get lost, which led to huge rotational errors. The important peculiarity of this dataset is a handheld camera and lots of motion which makes IMU data especially important here.

\begin{table*}[h]
\begin{center}
\centering
\caption{Results of SLAM algorithms tested on KITTI dataset.}
\resizebox{\textwidth}{!}{
\label{table:kitti_table}
\begin{tabular}{|c|c|c|c|c|c|c|c|c|c|c|c|c|c|c|c|c|} 
\hline
                      & \multicolumn{4}{c|}{\textbf{00}}                                                                                                                                                              & \multicolumn{4}{c|}{\textbf{02 }}                                                                                                                                                              & \multicolumn{4}{c|}{\textbf{05 }}                                                                                                                                                             & \multicolumn{4}{c|}{\textbf{06 }}                                                                                                                                                              \\ 
\cline{2-17}
\textbf{Framework}    & \multicolumn{2}{c|}{\textbf{position}}                                                        & \multicolumn{2}{c|}{\begin{tabular}[c]{@{}c@{}}\textbf{rotation} \\\end{tabular}}             & \multicolumn{2}{c|}{\begin{tabular}[c]{@{}c@{}}\textbf{position} \\\end{tabular}}              & \multicolumn{2}{c|}{\begin{tabular}[c]{@{}c@{}}\textbf{rotation} \\\end{tabular}}             & \multicolumn{2}{c|}{\textbf{position}}                                                        & \multicolumn{2}{c|}{\begin{tabular}[c]{@{}c@{}}\textbf{rotation} \\\end{tabular}}             & \multicolumn{2}{c|}{\textbf{position}}                                                        & \multicolumn{2}{c|}{\textbf{rotation}}                                                         \\ 
\cline{2-17}
                      & \textbf{max}                                  & \textbf{rms}                                  & \textbf{max}                                  & \textbf{rms}                                  & \textbf{max}                                   & \textbf{rms}                                  & \textbf{max}                                  & \textbf{rms}                                  & \textbf{max}                                  & \textbf{rms}                                  & \textbf{max}                                  & \textbf{rms}                                  & \textbf{max}                                  & \textbf{rms}                                  & \textbf{max}                                  & \textbf{rms}                                   \\ 
\hline
\textbf{ORB-SLAM2}    & 2.82                                          & \textcolor[rgb]{0.502,0,0.502}{\textbf{\textit{0.89}}}     & \textbf{\textcolor[rgb]{0,0.502,0}{6.67}} & \textcolor[rgb]{0.502,0,0.502}{\textbf{\textit{0.74}}}     & \textcolor[rgb]{0.502,0,0.502}{\textbf{\textit{15.76}}} & 6.60                                          & 4.37                                          & 1.47                                          & \textcolor[rgb]{0.502,0,0.502}{\textbf{\textit{0.94}}}     & \textcolor[rgb]{0.502,0,0.502}{\textbf{\textit{0.39}}}     & \textcolor[rgb]{0.502,0,0.502}{\textbf{\textit{2.16}}}     & \textcolor[rgb]{0.502,0,0.502}{\textbf{\textit{0.38}}}     & \textcolor[rgb]{0.502,0,0.502}{\textbf{\textit{1.14}}}     & \textcolor[rgb]{0.502,0,0.502}{\textbf{\textit{0.62}}}     & \textcolor[rgb]{0,0.502,0}{\textbf{0.76}} & \textcolor[rgb]{0,0.502,0}{\textbf{0.39}}  \\ 
\hline
\textbf{LDSO }        & 24.18                                         & 10.91                                         & 9.66                                          & 2.40                                          & 91.37                                          & 22.30                                         & 5.56                                          & 2.41                                          & 14.23                                         & 4.47                                          & 2.37                                          & 0.87                                          & 25.92                                         & 13.69                                         & 0.84                                          & 0.59                                           \\ 
\hline
\textbf{VINS-Fusion } & 13.80                                         & 5.20                                          & 7.93                                          & 3.22                                          & 47.90                                          & 18.17                                         & 9.31                                          & 4.78                                          & 16.78                                         & 4.79                                          & 7.49                                          & 3.02                                          & 7.14                                          & 2.50                                          & 6.93                                          & 4.81                                           \\ 
\hline
\textbf{OpenVSLAM }   & \textcolor[rgb]{0,0.502,0}{\textbf{2.41}} & \textcolor[rgb]{0,0.502,0}{\textbf{0.88}} & 6.94                                          & 0.75                                          & \textcolor[rgb]{0,0.502,0}{\textbf{12.88}}                                          & \textcolor[rgb]{0,0.502,0}{\textbf{5.53}} & \textbf{\textcolor[rgb]{0,0.502,0}{4.18}} & \textcolor[rgb]{0,0.502,0}{\textbf{1.20}} & 1.19                                          & 0.46                                          & \textcolor[rgb]{0,0.502,0}{\textbf{2.15}} & \textcolor[rgb]{0.502,0,0.502}{\textbf{\textit{0.38}}}     & 1.31                                          & 0.79                                          & 1.20                                          & 0.69                                           \\ 
\hline
\textbf{Basalt }      & 5.09                                          & 3.38                                          & 7.25                                          & 0.88                                          & 19.49                                          & 9.11                                          & 4.88                                          & 1.72                                          & 4.96                                          & 2.48                                          & 2.30                                          & 0.86                                          & 4.00                                          & 2.11                                          & 3.17                                          & 2.78                                           \\ 
\hline
\textbf{ORB-SLAM3 }   & \textcolor[rgb]{0.502,0,0.502}{\textbf{\textit{2.57}}}     & 0.90                                          & \textcolor[rgb]{0.502,0,0.502}{\textbf{\textit{6.85}}}     & \textcolor[rgb]{0,0.502,0}{\textbf{0.73}} & 16.27     & \textcolor[rgb]{0.502,0,0.502}{\textbf{\textit{6.35}}}     & \textcolor[rgb]{0.502,0,0.502}{\textbf{\textit{4.27}}}     & \textcolor[rgb]{0.502,0,0.502}{\textbf{\textit{1.43}}}     & \textcolor[rgb]{0,0.502,0}{\textbf{0.76}} & \textcolor[rgb]{0,0.502,0}{\textbf{0.36}} & 2.18                                          & \textbf{\textcolor[rgb]{0,0.502,0}{0.37}} & \textbf{\textcolor[rgb]{0,0.502,0}{0.67}} & \textbf{\textcolor[rgb]{0,0.502,0}{0.39}} & \textcolor[rgb]{0,0.502,0}{\textbf{0.76}}     & \textcolor[rgb]{0.502,0,0.502}{\textbf{\textit{0.41}}}      \\
\hline
\end{tabular}
}
\end{center}
\end{table*}

\begin{table*}[h]
\begin{center}
\centering
\caption{Results of SLAM algorithms tested on Open LORIS dataset.}
\label{table:open_loris_table}
\resizebox{\textwidth}{!}{
\begin{tabular}{|c|c|c|c|c|c|c|c|c|c|c|c|c|c|c|c|c|c|c|c|c|} 
\hline
\textbf{}                                                    & \multicolumn{4}{c|}{\textbf{office5 }}                                                                                                                                                          & \multicolumn{4}{c|}{\textbf{office7}}                                                                                                                                                           & \multicolumn{4}{c|}{\textbf{cafe1 }}                                                                                                                                                          & \multicolumn{4}{c|}{\textbf{cafe2 }}                                                                                                                                                        & \multicolumn{4}{c|}{\textbf{ home1 }}                                                                                                                                                         \\ 
\cline{2-21}
\textbf{Framework}                                           & \multicolumn{2}{c|}{\textbf{position }}                                                       & \multicolumn{2}{c|}{\textbf{rotation }}                                                       & \multicolumn{2}{c|}{\textbf{position }}                                                       & \multicolumn{2}{c|}{\textbf{rotation }}                                                       & \multicolumn{2}{c|}{\textbf{position }}                                                       & \multicolumn{2}{c|}{\textbf{rotation }}                                                       & \multicolumn{2}{c|}{\textbf{position }}                                                       & \multicolumn{2}{c|}{\textbf{rotation }}                                                       & \multicolumn{2}{c|}{\textbf{position }}                                                       & \multicolumn{2}{c|}{\textbf{rotation }}                                                        \\ 
\cline{2-21}
                                                             & \textbf{max}                                  & \textbf{rms}                                  & \textbf{max}                                  & \textbf{rms}                                  & \textbf{max}                                  & \textbf{rms}                                  & \textbf{max}                                  & \textbf{rms}                                  & \textbf{max}                                  & \textbf{rms}                                  & \textbf{max}                                  & \textbf{rms}                                  & \textbf{max}                                  & \textbf{rms}                                  & \textbf{max}                                  & \textbf{rms}                                  & \textbf{max}                                  & \textbf{rms}                                  & \textbf{max}                                  & \textbf{rms}                                   \\ 
\hline
\textbf{ORB-SLAM2}                                           &  
0.30     & \textcolor[rgb]{0,0.502,0}{\textbf{0.21}} & 7.45                                          & 2.97     & \textcolor[rgb]{0,0.502,0}{\textbf{0.09}} & \textcolor[rgb]{0,0.502,0}{\textbf{0.04}} & 2.88    & \textcolor[rgb]{0,0.502,0}{\textbf{1.45}} &
0.36     & 0.21     & 5.04     & 2.72     & 
X & X & X & X & 
X & X & X & X \\ 
\hline
\begin{tabular}[c]{@{}c@{}}\textbf{VINS-Mono}\\\end{tabular} &  0.32                                          & 0.22     & 5.83     & 3.38                                          & 0.40                                          & 0.12                                          & 3.75                                          & 2.42 &                                         1.31                                          & 0.42                                          & 5.40                                          & 3.47                                          & 0.75                                          & \textcolor[rgb]{0,0.502,0}{\textbf{0.15}}     & \textcolor[rgb]{0,0.502,0}{\textbf{4.27}} & \textcolor[rgb]{0,0.502,0}{\textbf{2.79}}      &
1.09                                          & 0.55                                          & 16.01                                        & 8.86\\ 
\hline
\begin{tabular}[c]{@{}c@{}}\textbf{OpenVSLAM}\\\end{tabular} & 
\textcolor[rgb]{0,0.502,0}{\textbf{0.28}} & 0.23                                          & \textcolor[rgb]{0,0.502,0}{\textbf{5.12}} & 1.26 & 0.10     & 0.05     & \textcolor[rgb]{0,0.502,0}{\textbf{2.86}} & 1.55     &
\textcolor[rgb]{0,0.502,0}{\textbf{0.14}} & \textcolor[rgb]{0,0.502,0}{\textbf{0.10}} & \textcolor[rgb]{0,0.502,0}{\textbf{3.35}} & \textcolor[rgb]{0,0.502,0}{\textbf{0.77}} & \textcolor[rgb]{0,0.502,0}{\textbf{0.42}}     & 0.22                                          & 6.37                                          & 4.35                                          & 
\textcolor[rgb]{0,0.502,0}{\textbf{0.50}}      & \textcolor[rgb]{0,0.502,0}{\textbf{0.36}}     & \textcolor[rgb]{0,0.502,0}{\textbf{9.36}}     & \textcolor[rgb]{0,0.502,0}{\textbf{2.30}}     \\
\hline
\end{tabular}
}
\end{center}
\end{table*}

Table \ref{table:kitti_table} represents errors on KITTI dataset. Only the algorithms that could work with stereo data are provided. ORB-SLAM2 and OpenVSLAM show pretty equal results with low errors (RMSE $<$ 1 meter) on the majority of sequences. VINS Fusion and Basalt have several times worse results. Meanwhile, LDSO suffers much on KITTI sequences. As can be noticed, all methods have difficulties with sequence 02, which has an irregular shape and many smooth curves.

\begin{figure*}[h]
    \centering
    \includegraphics[width=1.0\textwidth]{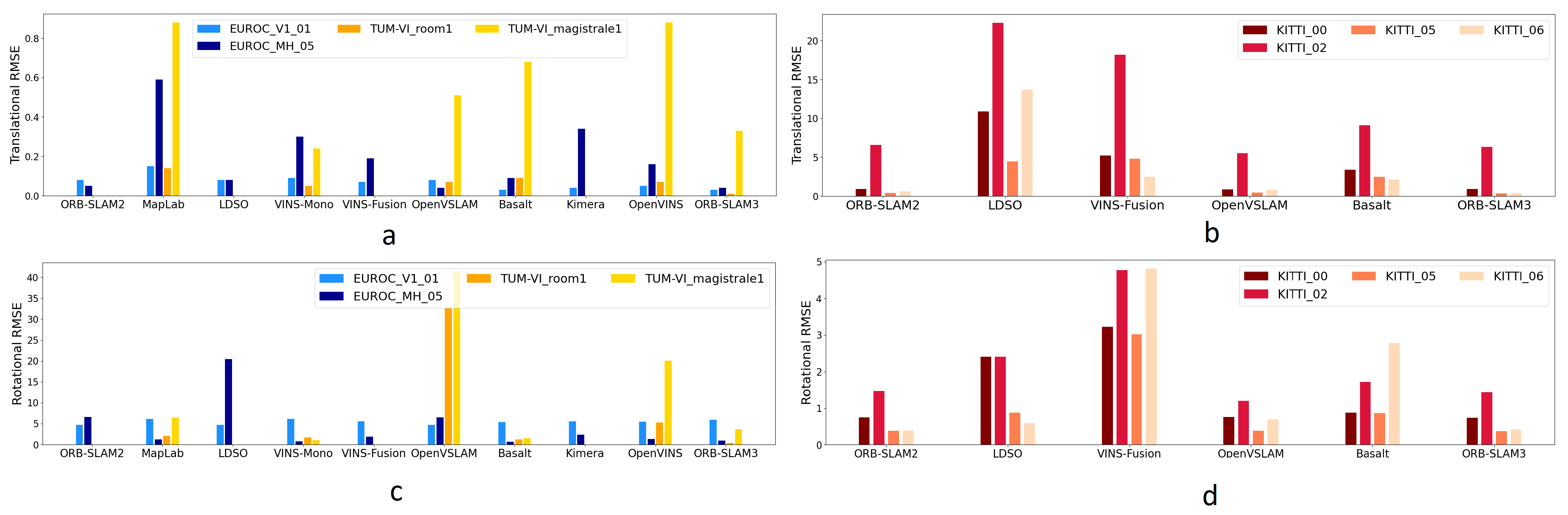}
    \caption{Comparison of translational (a-b) \& rotational (c-d) root mean squared errors on EuRoC, TUM VI and KITTI datasets.}
    \label{fig:metrics_tum_kitti_euroc}
\end{figure*}


\begin{table*}[h]
\centering
\caption{CPU Load and Memory Usage.}
\label{table:cpu_ram_table}
\begin{adjustbox}{width=1\textwidth}
\begin{tabular}{|c|c|c|c|c|c|c|c|} 
\hline
\textbf{Framework} & \textbf{Dataset} & \multicolumn{3}{c|}{\begin{tabular}[c]{@{}c@{}}\textbf{Memory RAM (Mb)} \\\end{tabular}} & \multicolumn{3}{c|}{\textbf{CPU Load (\%)}} \\ 
\cline{3-8}
 &  & \textbf{Average} & \textbf{Max} & \textbf{Std} & \textbf{Average} & \textbf{Max} & \textbf{Std} \\ 
\hline
\textbf{Maplab} &  & 526.5 & 721.7 & 36.1 & \textcolor[rgb]{0.502,0,0.502}{\textbf{\textit{85.2}}} & 523.2 & \textbf{\textcolor[rgb]{0.502,0,0.502}{\textit{34.9}}} \\ 
\cline{1-1}\cline{3-8}
\textbf{VINS-Mono} & TUM VI corridor & \textbf{\textcolor[rgb]{0,0.502,0}{87.0}} & \textbf{\textcolor[rgb]{0,0.502,0}{92.1}} & \textbf{\textcolor[rgb]{0,0.502,0}{1.1}} & \textbf{\textcolor[rgb]{0,0.502,0}{68.5}} & \textbf{\textcolor[rgb]{0,0.502,0}{118.8}} & \textbf{\textcolor[rgb]{0,0.502,0}{36.2}} \\ 
\cline{1-1}\cline{3-8}
\textbf{Basalt} &  & \textbf{\textcolor[rgb]{0.502,0,0.502}{\textit{107.8}}} & \textbf{\textcolor[rgb]{0.502,0,0.502}{\textit{109.2}}} & \textcolor[rgb]{0.502,0,0.502}{\textbf{\textit{1.3}}} & 418.3 & 560.1 & 56.9 \\ 
\cline{1-1}\cline{3-8}
\textbf{OpenVINS} &  & 1686.2 & 3329.1 & 952.5 & \textcolor[rgb]{0.502,0,0.502}{\textbf{\textit{85.2}}} & \textcolor[rgb]{0.502,0,0.502}{\textbf{\textit{523}}} & \textcolor[rgb]{0.502,0,0.502}{\textbf{\textit{34.9}}} \\ 
\hline
\textbf{Maplab} &  & \textcolor[rgb]{0.502,0,0.502}{\textbf{1587.5}} & 4587.4 & \textbf{\textcolor[rgb]{0.502,0,0.502}{\textit{682.6}}} & \textcolor[rgb]{0.502,0,0.502}{\textbf{\textit{160.6}}} & 792.8 & 93.2 \\ 
\cline{1-1}\cline{3-8}
\textbf{VINS-Mono} &  & 2851.0 & \textcolor[rgb]{0.502,0,0.502}{\textbf{\textit{4524.2}}} & 1285.9 & 182.5 & 306.7 & 94.2 \\ 
\cline{1-1}\cline{3-8}
\textbf{OpenVSLAM} & TUM VI outdoors & 2639.6 & 4363.4 & 1071.3 & 230.6 & \textcolor[rgb]{0.502,0,0.502}{\textbf{\textit{310.5}}} & \textcolor[rgb]{0.502,0,0.502}{\textbf{\textit{38.8}}} \\ 
\cline{1-1}\cline{3-8}
\textbf{Basalt} &  & \textbf{\textcolor[rgb]{0,0.502,0}{136.5}} & \textbf{\textcolor[rgb]{0,0.502,0}{161.5}} & \textbf{\textcolor[rgb]{0,0.502,0}{8.0}} & 537.7 & 676.9 & 57.9 \\ 
\cline{1-1}\cline{3-8}
\textbf{OpenVINS} &  & 6490.8 & 12819.5 & 3694.5 & \textbf{\textcolor[rgb]{0,0.502,0}{100.5}} & \textbf{\textcolor[rgb]{0,0.502,0}{147.8}} & \textbf{\textcolor[rgb]{0,0.502,0}{14.3}} \\ 
\hline
\textbf{ORB-SLAM2} &  & \textcolor[rgb]{0.502,0,0.502}{\textbf{\textit{705.9}}} & \textcolor[rgb]{0.502,0,0.502}{\textbf{\textit{869.9}}} & \textcolor[rgb]{0.502,0,0.502}{\textbf{\textit{102.7}}} & 187.2 & \textcolor[rgb]{0.502,0,0.502}{\textbf{\textit{249.8}}} & \textcolor[rgb]{0.502,0,0.502}{\textbf{\textit{37.0}}} \\ 
\cline{1-1}\cline{3-8}
\textbf{LDSO} & Euroc MH\_05 & 946.3 & 1082.1 & 102.9 & \textcolor[rgb]{0.502,0,0.502}{\textbf{\textit{160.1}}} & 263.9 & 50.6 \\ 
\cline{1-1}\cline{3-8}
\textbf{Kimera} &  & \textbf{\textcolor[rgb]{0,0.502,0}{250.5}} & \textbf{\textcolor[rgb]{0,0.502,0}{273.6}} & \textbf{\textcolor[rgb]{0,0.502,0}{46.4}} & \textbf{\textcolor[rgb]{0,0.502,0}{80.5}} & \textbf{\textcolor[rgb]{0,0.502,0}{120.8}} & \textbf{\textcolor[rgb]{0,0.502,0}{17.6}} \\ 
\hline
\textbf{DRE-SLAM} & DRE dataset hd2 & \textbf{\textcolor[rgb]{0,0.502,0}{5010.7}} & \textbf{\textcolor[rgb]{0,0.502,0}{6288.1}} & \textbf{\textcolor[rgb]{0,0.502,0}{1777.0}} & \textbf{\textcolor[rgb]{0,0.502,0}{200.0}} & \textbf{\textcolor[rgb]{0,0.502,0}{303.5}} & \textbf{\textcolor[rgb]{0,0.502,0}{77.3}} \\
\hline
\end{tabular}
\end{adjustbox}
\end{table*}

The summarized results of root mean squared errors for translation (in meters) and rotation (in degrees) obtained on three main datasets (TUM VI, EuRoC, KITTI) in 
Fig. \ref{fig:metrics_tum_kitti_euroc} with different subset sequences. Histograms with the KITTI dataset are highlighted separately as they are based on traffic roads and one may be curious about the performance of different approaches in outdoor dynamic conditions. In terms of translation and rotation errors, ORB-SLAM 2/3 shows its superiority for all the sequences from each dataset. MapLab, OpenVSLAM, Basalt, and OpenVINS show mediocre results on TUM VI Magistrale 1 sequence, but OpenVSLAM has additional problems in rotation. VINS-Fusion and LDSO both obtain high errors on KITTI sequences, which impugns the usage of these approaches on public roads. Rotational errors on EUROC data are on a similar level of magnitude for all the algorithms except LDSO. This could be explained by the nature of data from a micro aerial vehicle. ORB-SLAM3, Basalt, Vins-Mono, and MapLab have the lowest rotational errors on the TUM VI sequences. Whereas OpenVSLAM is the only visual method on TUM VI that shows the highest error.

Experiments on the Open LORIS dataset (Table \ref{table:open_loris_table}) aim to compare successful algorithms from previous runs that use data from different types of sensors. There are monocular rgb-d ORB-SLAM2, stereo VINS-Mono with IMU data, and OpenVSLAM with images from a stereo fish-eye camera but without IMU. The table shows that in general, all three algorithms have similar performance on several sequences. Looking at the details, it can be noticed that on the most crowded and dynamic sequences (cafe2 and home1), ORB-SLAM2 is losing trajectory at some moment. Moreover, monocular VINS mono overall has higher errors than stereo fish-eye OpenVSLAM. LDSO could not initialise these data. Also, the wheel encoder data is in an inappropriate data format for the DRE-SLAM. 

In Table \ref{table:dre} the results of ORB-SLAM2 (mono and rgb-d versions) with DRE-SLAM on the DRE dataset with dynamic content are illustrated. It can be clearly seen that DRE SLAM \cite{dre} gives the best results both on low dynamic and high dynamic sequences in comparison with ORB-SLAM2.

Table \ref{table:cpu_ram_table} indicates CPU load and RAM memory usage. All algorithms were tested on the system with Intel Core i7 8565U CPU and 16 GB of RAM. It is worth mentioning that besides external modules of some approaches working in parallel the amount of required memory depends on many other aspects. A number of keyframes, amount of tracked visual features per keyframe,  the type of a feature, and descriptor. The denser the map the more memory and computational time are needed to solve the problem. If the back-end of the SLAM algorithm uses all the measurements in batch mode: Open VSLAM, ORB-SLAM 2/3, LDSO  with g2o \cite{Kummerle2011} solver or VINS-Mono, VINS-Fusion, Maplab, DRE with Ceres \cite{CERES} solver, obviously, computational resources are limited, but this paradigm provides the best accuracy if not a significant amount of outliers present. However, the Bayes tree structure of the factor graph proposed in iSAM/iSAM2 \cite{iSAM2} used in Kimera allows to significantly save resources due to incremental updates of Jacobian matrices with upcoming measurements. Only remote loop closures (between initial and last states) can nullify the advantage of a tree-based structure as all the poses have to be recalculated. The cheapest way to update trajectory and map is a fixed-lag smoothing approach with marginalization \cite{marginalisation} used in Basalt which is a bit similar to filtering used in Open VINS. Unfortunately, it does not allow us to use all the measurements and might cause problems with remote loop closures. Recent algorithms use to solve a full problem with all measurements in a parallel thread for saving resources while other threads are busy with current estimation. As table \ref{table:cpu_ram_table} illustrates, VINS-Mono demands fewer resources than others on the TUM VI corridor (indoor) environment but rather mediocre results on the outdoor sequence. Basalt shows a small need for memory both on indoor and outdoor sequences. OpenVINS consumes the smallest amount of CPU resources on TUM VI outdoors. All other algorithms on the outdoors sequence show significant demand for memory. The reason for that is the length of the sequence, leading to a huge amount of visual features. Kimera shows itself to be quite sufficient in memory consumption and CPU load on Euroc MH\_05 sequence. Results of ORB-SLAM2 and LDSO are comparable and quite demanding both in memory and CPU load. DRE-SLAM expectedly demands lots of resources even for a relatively short sequence. The possible reason is the use of several modules for a bunch of tasks. 

\section{Conclusions and recommendations}
After a deep comparison of different algorithms, it could be concluded that ORB-SLAM2 and ORB-SLAM3 on average are the most reliable approaches for trajectory estimation. They show good results in different environments, sensor setups and even with dynamic objects. The reasons for that are a quality implementation and both fast and reliable visual features and the back-end part of the algorithm.

From a practical point of view, there is no perfect open-source out-of-the-box VIO/Visual SLAM solution for different environments and conditions. It is always needed to tune algorithm parameters for the particular task, preprocess data, and solve dependency and performance issues.

Visual-inertial approaches: VINS-Mono, Basalt and OpenVINS benefit from using IMU data and show competitive results for versatile datasets. The drawback of visual-inertial fusion is the need for additional data sources, however, the majority of datasets provide this information. Meanwhile, one can mention that visual-only ORB-SLAM2, which was published five years ago, still can fight for 1-2 place. The same applies to the similar but refined OpenVSLAM, which got one of the best results on many datasets except TUM VI. Kimera represents a nice modular concept with great opportunities but low usability. Thus, it was hard to launch it on other datasets except for EuRoC. The only dense approach (LDSO) that had been tested showed far worse results compared to sparse algorithms. The modern ORB-SLAM3 shows its superiority in different setups and seems to be a perspective option to continue experiments. It tends to be a leader for the majority of datasets because of the welding map merging method \cite{orb3} and novel place recognition technique. Finally, the full Bundle Adjustment problem is being solved \cite{BundleAdjustement}. Unfortunately, many dynamic SLAM approaches failed to run because of various issues. DRE-SLAM showed perfect results on dynamic data with the help of wheel encoders. But the problem is in the lack of wheel encoder data in datasets and it is an additional constraint to the setup during real-life data recording. The overall conclusion is that for different data and environments various approaches could show qualitative results and there is no striking leader. The most important is a correct experimental setup and the concept of limitations. 

Another important point is that many SLAM solutions which have promising results in the paper but completely unusable GitHub repositories with a lack of support and examples only on the benchmark datasets. This refers to the reproducibility crisis which was stated in the introduction part of the paper. 

Support and relevance of the implementation are important, as well as detailed instructions by the authors. Quickly updating packages cause dependency and repeatability issues even for a three-year-old solution.

The same is true for datasets. The absence of a complete dataset that could allow comparing different types of algorithms makes it difficult to understand which approach is the best one. Moreover, there is a need to unify data organisation both in datasets and in SLAM algorithms. Different formats of the ground truth data add a mess to work. For example, the KITTI format without timestamps makes it unusable for comparison of trajectories in other formats because of timestamps absence. Nevertheless, there is a significant choice of datasets suitable for comparing distinctive types of algorithms or specific environmental conditions.  

It could be concluded that a user-friendly open-source slam solution is rather rare. One of them was OpenVSLAM. But during the paper writing, it has been deleted from the GitHub\footnote{ \url{https://github.com/xdspacelab/openvslam/wiki/Termination-of-the-release}} due to the concerns on similarities of the source code to the ORB-SLAM2. Indeed, it can be noted that the resulting trajectories and errors are very similar in these two algorithms. Even though the National Institute of Advanced Industrial Science and Technology did not find the copyright infringement incidences, they decided to terminate the release of OpenVSLAM to avoid any risk of possible copyright issues. 

Finally, this would be very convenient if every research work or SLAM approach had a Docker container to run and check the results described in the proposed method as it might solve reproducibility issues. even though the performance of an algorithm depends on computational resources, the Docker image allows using the same versions of the software so the difference will be only at a hardware level. To prove the practical consistency of an algorithm, it would be helpful if the latter had a ROS wrapper to check the proposed method on a real robot as a majority of the robotics community uses ROS. Also, this would be meaningful if robotics datasets had as many sensors as possible in order to check more approaches for a better understanding of benefits and disadvantages. Some methods could provide significantly better results on a dataset if there are sensor measurements needed for a visual SLAM approach.

It is also important to highlight several possible directions for future research in the area. For real-life applications, it could be reasonable to test algorithms on hardware-constrained microcomputing platforms. Another important aspect is the reliability of approaches for very long traversals, such as rides of self-driving cars. Lastly, it could be very beneficial to test the robustness of algorithms to outliers through a comparison of solvers and front-end parts of the algorithms. 

\section{Declarations}
\textbf{Funding:}
\\The authors declare that no funds, grants, or other support were received during the preparation of this manuscript.
\\\textbf{Conflicts of interest:}
\\The authors have no relevant financial or non-financial interests to disclose, as well as no competing interests to declare that are relevant to the content of this article.
\\\textbf{Authors' contributions:}
\\All authors contributed to the study conception and design. Material preparation, data collection and analysis were performed by Dinar Sharafutdinov, Mark Griguletskii, Pavel Kopanev, Mikhail Kurenkov and Aleksey Burkov. The review, text editing and article structure was made by Dzmitry Tsetserukou, Aleksei Gonnochenko and Gonzalo Ferrer.



\bibliographystyle{spmpsci}
\bibliography{references.bib}{}

\end{document}